\documentclass[10pt,journal]{IEEEtran}

\ifCLASSOPTIONcompsoc
  \usepackage[nocompress]{cite}
\else
  \usepackage{cite}
\fi

\usepackage{graphicx}
\usepackage{times}
\usepackage{epsfig}
\usepackage{graphicx}
\usepackage{amsmath}
\usepackage{amssymb}
\usepackage{multicol}
\usepackage{balance}
\usepackage{multirow}
\usepackage{float}
\usepackage{caption}
\usepackage{ctable}
\usepackage{array}
\usepackage{makecell}
\usepackage{subfig}
\usepackage{tabularx}
\usepackage{bbding}
\usepackage{pifont}
\usepackage{url}
\usepackage{times}
\usepackage{epsfig}
\usepackage{graphicx}
\usepackage{amsmath}
\usepackage{amssymb}
\usepackage{multirow}
\usepackage{colortbl}
\usepackage{color}
\usepackage{threeparttable}
\usepackage{booktabs}
\usepackage{adjustbox}
\usepackage{subfig}
\usepackage{caption}
\usepackage[misc]{ifsym} 
\usepackage{xfrac}
\usepackage{tabto,enumitem}
\usepackage[misc]{ifsym} 
\usepackage[linesnumbered,lined,boxed,commentsnumbered,ruled]{algorithm2e}
\usepackage[pagebackref,breaklinks=true,colorlinks,citecolor=blue,urlcolor=blue,linkcolor=blue,bookmarks=false]{hyperref}
\usepackage{xspace}
\newcommand*{\eg}{e.g.\@\xspace}
\newcommand*{\ie}{i.e.\@\xspace}

\begin{document}
%
\title{Local-to-Global Information Communication for Real-Time Semantic Segmentation Network Search}
\newcommand{\guangliang}[1]{{\color{blue}(guangliang: {#1})}} 
\author{Guangliang Cheng$^*$, Peng Sun$^*$, Ting-Bing Xu, Shuchang Lyu and Peiwen Lin$^\textrm{\Letter}$         
        \thanks {Guangliang Cheng is with SenseTime Research Group of Beijing, and Department of Computer Science, in the University of Liverpool.}
        \thanks{Peng Sun is with Bytedance Data Group}
        \thanks{Ting-Bing Xu and Shuchang Lyu are with Beihang University}
        \thanks{Peiwen Lin is with SenseTime Research Group of Beijing}
        \thanks{$^*$ The first two authors contributed equally to this paper.}
        \thanks{$^\textrm{\Letter}$ Corresponding author: Peiwen Lin}                 
}

\markboth{Journal of \LaTeX\ Class Files,~Vol.~XX, No.~X, XX~XXXX}%
{Shell \MakeLowercase{\textit{et al.}}: Bare Demo of IEEEtran.cls for IEEE Journals}
%



\maketitle



%
\IEEEpeerreviewmaketitle


\begin{abstract}
  %
  Neural Architecture Search (NAS) has shown great potentials in automatically designing neural network architectures for real-time semantic segmentation. Unlike previous works that utilize a simplified search space with cell-sharing way, we introduce a new search space where a lightweight model can be more effectively searched by replacing the cell-sharing manner with cell-independent one. Based on this, the communication of local to global information is achieved through two well-designed modules. For local information exchange, a graph convolutional network (GCN) guided module is seamlessly integrated as a communication deliver between cells. For global information aggregation, we propose a novel dense-connected fusion module (cell) which aggregates long-range multi-level features in the network automatically. In addition, a latency-oriented constraint is endowed into the search process to balance the accuracy and latency. We name the proposed framework as Local-to-Global Information Communication Network Search (LGCNet). Extensive experiments on Cityscapes and CamVid datasets demonstrate that LGCNet achieves the new state-of-the-art trade-off between accuracy and speed. In particular, on Cityscapes dataset, LGCNet achieves the new best performance of 74.0\% mIoU with the speed of 115.2 FPS on Titan Xp.

\end{abstract}

\begin{IEEEkeywords}
  Neural Architecture Search, Real-Time Semantic Segmentation, Graph Convolutional Network.
\end{IEEEkeywords}
%
%
%
%

 




  \section{Introduction}
  \label{intro}

  \IEEEPARstart{S}{emantic} segmentation~\cite{long2015fully,zhao2017pyramid,chen2018deeplab,DBLP:journals/corr/ChenPSA17,DBLP:journals/tip/LiZCYTZX21, DBLP:journals/tip/LiLYZCYTL21} has been a fundamental vision task that aims at predicting pixel-level semantic categories for images. With the recent advances in deep learning technology~\cite{corr_SimonyanZ14a,conf_cvpr_HeZRS16,journals_corr_HuangLW16a,DBLP:conf/cvpr/Chollet17,DBLP:journals/corr/abs-2107-13155, DBLP:conf/eccv/LiLZCSLTT20}, many works focus on the sophisticated model design regarding depth, width and attention mechanism to pursue higher accuracy, which involves many time-consuming operations. Although these methods achieve impressive results on the public semantic segmentation benchmarks~\cite{cordts2016cityscapes,everingham2015pascal,Camvid}, they are difficult to be deployed on the resource-constrained applications, such as the auto-driving vehicles and the navigation robots, which require high computational efficiency without incurring accuracy drop. To overcome this problem, researchers have designed some low-computation CNN models to harvest the satisfactory segmentation accuracy. ICNet, ENet and SegNet~\cite{zhao2018icnet,badrinarayanan2017segnet,paszke2016enet} reduce the computational cost by incorporating some specifically designed modules with reduced input size or filter numbers. BiSeNet~\cite{DBLP:conf/eccv/YuWPGYS18,yu2021bisenet} decouples the context and spatial information with bilateral path, and then fuses them to achieve a satisfactory trade-off between accuracy and latency. DFANet~\cite{li2019dfanet} utilizes lightweight depth-wise separable convolutions and remedies its accuracy drop by incorporating an aggregation module. Although achieving remarkable results, these human-designed strategies usually require expertise in architecture design through enormous trial and error to carefully balance the accuracy and resource-efficiency.
  Such human-designed process need to be redone when the hardware setting changes, which further increases the difficulty in applying to the actual applications.
  
  \begin{figure}[t]
  \centering
  \includegraphics[width=1.05\columnwidth]{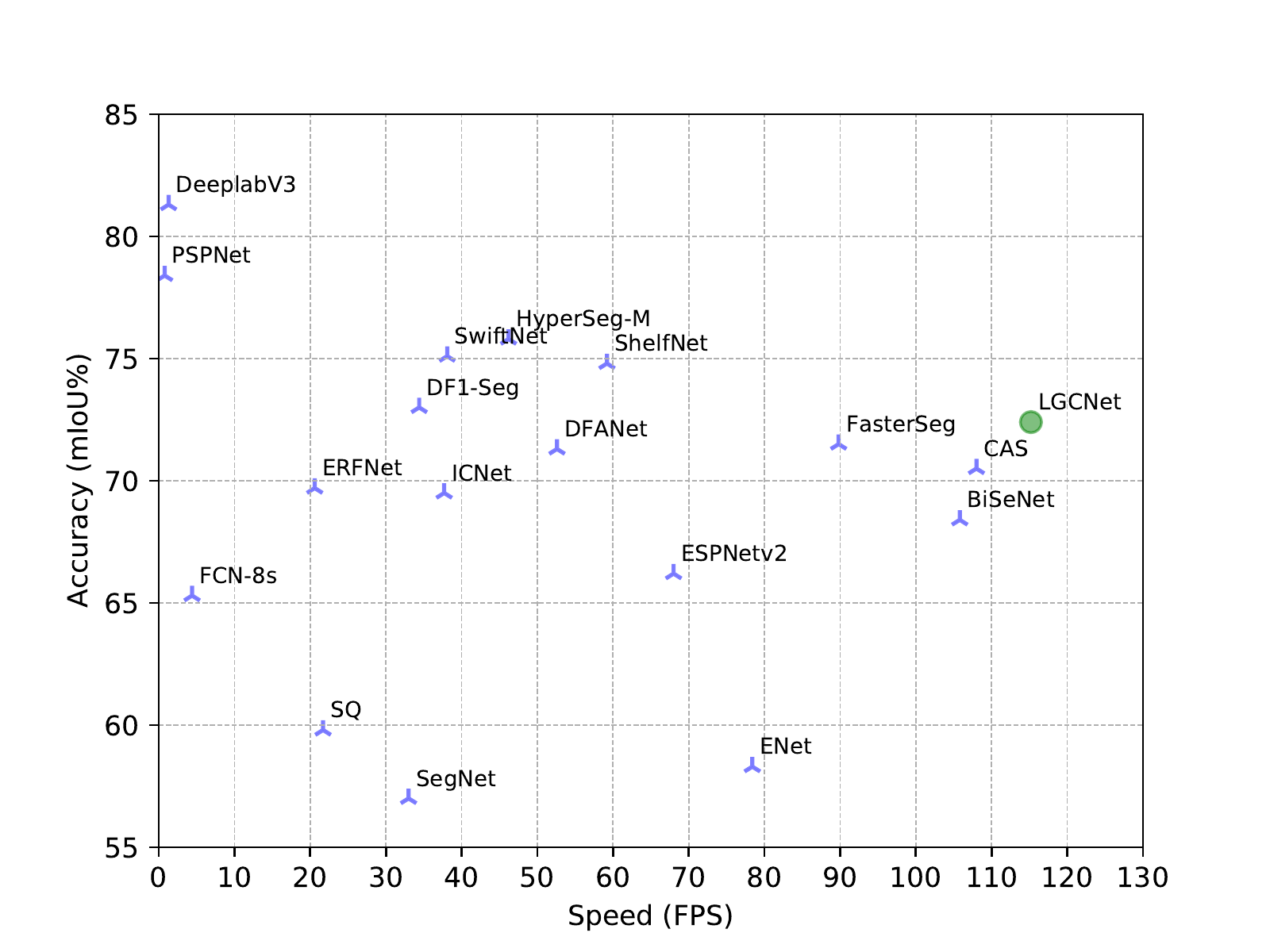}
  \caption{The inference speed (FPS) and mIoU performance among different networks on the Cityscapes test set. Our LGCNet achieves the state-of-the-art trade-off between the speed and performance.}
  
  \label{fig:plot}
  \end{figure}
  
  Different from manually designed architectures, network architecture search methods have drawn extensive attention \cite{DBLP:conf/iclr/LiuSY19,zoph2018learning,negrinho2017deeparchitect,krause2017dynamic,DBLP:conf/icml/PhamGZLD18,cai2018proxylessnas,DBLP:conf/iclr/XieZLL19,nekrasov2019fast,nirkin2021hyperseg} and achieved remarkable performance. They mainly focus on the block-based (also called as cell) search for the backbone and utilize the multi-scale module such as ASPP \cite{DBLP:conf/eccv/ChenZPSA18} to fuse the global information. For the building block, some works directly design the search space on the targeted platform. CAS~\cite{zhang2019customizable} searches for two cell types (normal and reduction cell) and then stacks the identical cells repeatedly to form a network. FBNet~\cite{wu2019fbnet} and SqueezeNet~\cite{2019_SqueezeNAS} search the hyper-parameters including number of blocks, channel numbers of each layer for the network based on the efficient blocks in human-designed network such as ResNet and MobileNet \cite{ conf_cvpr_HeZRS16, howard2017mobilenets}. For multi-scale module, AutoRTNet \cite{sun2021real} and CAS \cite{zhang2019customizable} automatically aggregate features at different levels with a multi-scale fusion cell, and \cite{nekrasov2019fast} utilizes a recurrent neural network (i.e. controller) to decide which layer and what kind of operations will be employed.
  

  
\begin{figure}[t]
  \centering
  \includegraphics[width=8cm]{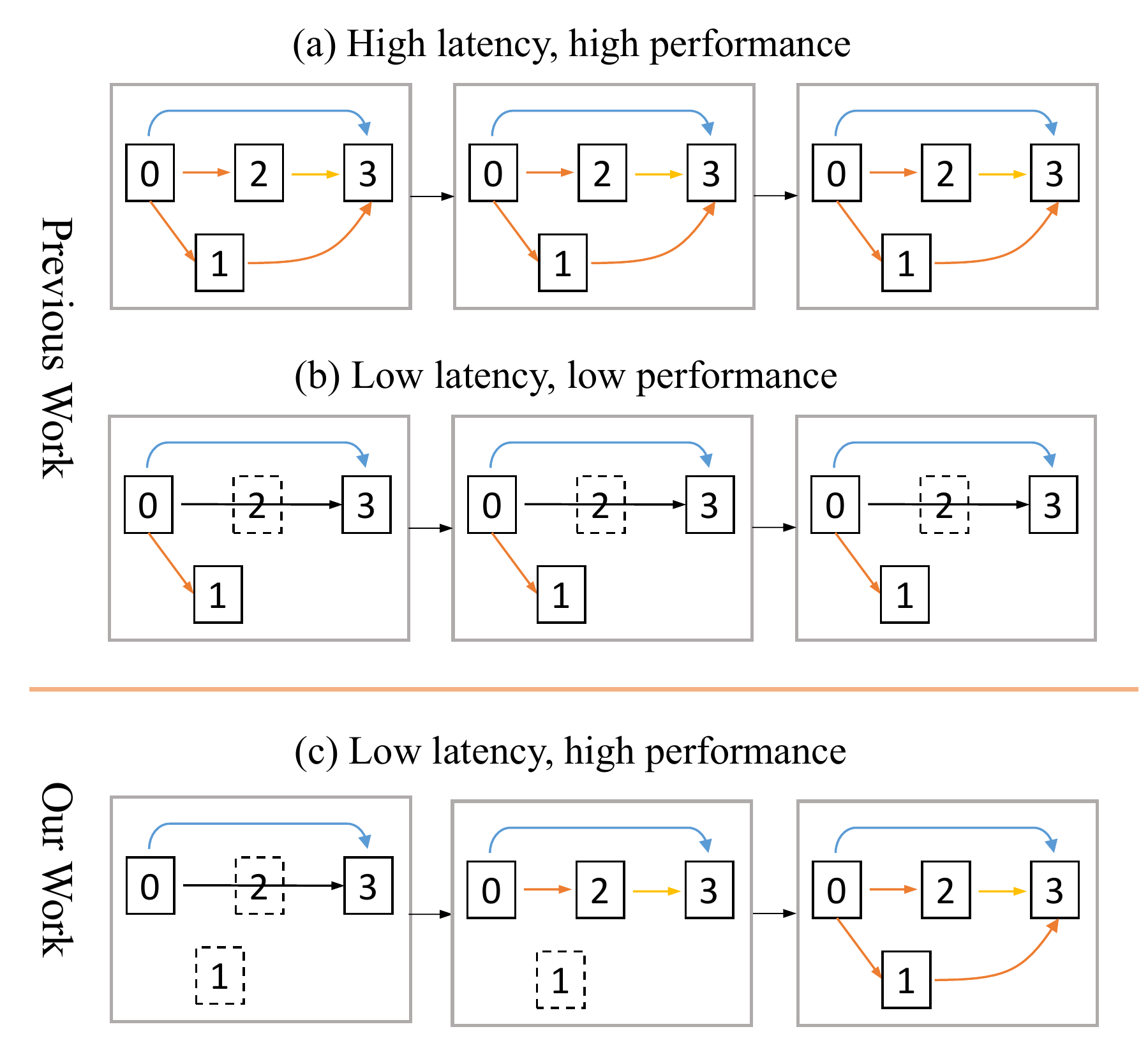}
  \caption{(a) The network stacked with complicated cells results in high latency and high performance. (b) The network stacked with simple cells leads to low latency and low performance. (c) The cell diversity strategy, \ie,~each cell possesses its own independent structure, can flexibly construct the high accuracy lightweight network.}
  
  \label{fig:flowchart}
  \end{figure}

  Although the searched building block and multi-scale cell have achieved satisfactory performance with above methods, some indispensable aspects for a remarkable real-time segmentation network are ignored: 
  
  1) Difficult to achieve a good trade-off between latency and accuracy due to the limited cell diversity with the identical cell. As shown in Figure~\ref{fig:flowchart} (a), the cell is prone to learn a complicated structure to achieve high performance without any resource constraint, and the network stacked with it will result in high latency. When a low-computation constraint is applied, the cell structure tends to be over-simplified as shown in Figure~\ref{fig:flowchart} (b), which may not achieve satisfactory performance. We thus modify the cell setting from cell-sharing manner to cell-independent one, which can be flexibly stacked to form a lightweight network with cell diversity as shown in Figure~\ref{fig:flowchart} (c).
  
  2) Lack of local information exchange. Above methods only consider the global information fusion using the multi-scale module during the search progress, while the local information exchange between adjacent cells are very important to achieve a good trade-off between latency and accuracy because different cells can be treated as multiple agencies, whose achievement of social welfare may require information exchange between them inspired by~\cite{minsky1988society}.
  
  3) Global information fusion. CAS~\cite{zhang2019customizable} and~\cite{nekrasov2019fast} search a multi-scale module in which only the short-range features from stride=8 to stride=16 are taken into account to reduce the inference time cost, but it would be helpful to improve the accuracy if the lower level information in stride=4 could be incorporated since some boundary clues from this feature are essential for achieving more fine-grained segmentation results.
  
  Based on the independent cell mechanism, it's worth further exploring how to effectively establish the local-to-global information communication among cells of the whole network during the search process. With this aim in mind, in this article we present a new lightweight segmentation network search method by integrating the local information exchange and global information fusion together from the local and global aspects. First, to address the local exchange, we utilize a Graph Convolutional Network~\cite{kipf2016semi} guided module (GGM) as the local information exchange deliverer among cells, through which the information of each cell can be propagated to the next adjacent cell. Second, a dense-connected fusion cell is proposed to perform global information aggregation, in which the long-range multi-level features (low-level spatial details and high-level semantic context) in the network can be effectively exploited and fused. A latency-oriented constraint for target computing platform is embedded into the search process, thus the searched model can achieve better trade-off between the accuracy and latency.
  
  \par To verify the effectiveness of the proposed search method, extensive experiments and detailed analysis are provided on two public segmentation datasets, i.e., the standard Cityscapes \cite{cordts2016cityscapes} and CamVid \cite{Camvid} benchmarks. The experimental results show that our method achieves much better performance and faster inference speed than GAS~\cite{gas}, which is a prior state-of-the-art trade-off between accuracy and speed. Compared to other real-time methods, our method also locates in the top-right area in Figure~\ref{fig:plot}, which indicates that our method obtains the new state-of-the-art trade-off between accuracy and latency.

\par This article is an extension of our conference version~\cite{gas}, and the major contributions can be summarized as follows.
\begin{itemize}
  \item We extend previous segmentation network architecture search from only graph-guided local information exchange between adjacent cells to the whole information communication of local perception and global fusion.
  \item The global information aggregation is implemented via the dense-connected fusion cell, which aggregates multi-level features automatically to effectively fuse the low-level spatial details and high-level semantic context.  
  \item More detailed algorithm descriptions, deeper analyses, and more comparison experiments are presented in this paper to demonstrate the effectiveness of the proposed method for the lightweight segmentation network architecture search.
  \item The lightweight segmentation model via the proposed search method is customizable in practical applications. Notably, it achieves 74.0\% mIoU on the Cityscapes test set and 115.2 FPS on NVIDIA Titan Xp for one 769$\times$1537 image.
\end{itemize}
\par \noindent The remainder of this article is organized as follows. Section~II reviews the related works; Section~III describes the details of the proposed method; Section~IV presents experimental results and discussions, and Section~V draws concluding remarks.


  \begin{figure*}[t]
  \begin{center}
  \includegraphics[width=6.8in]{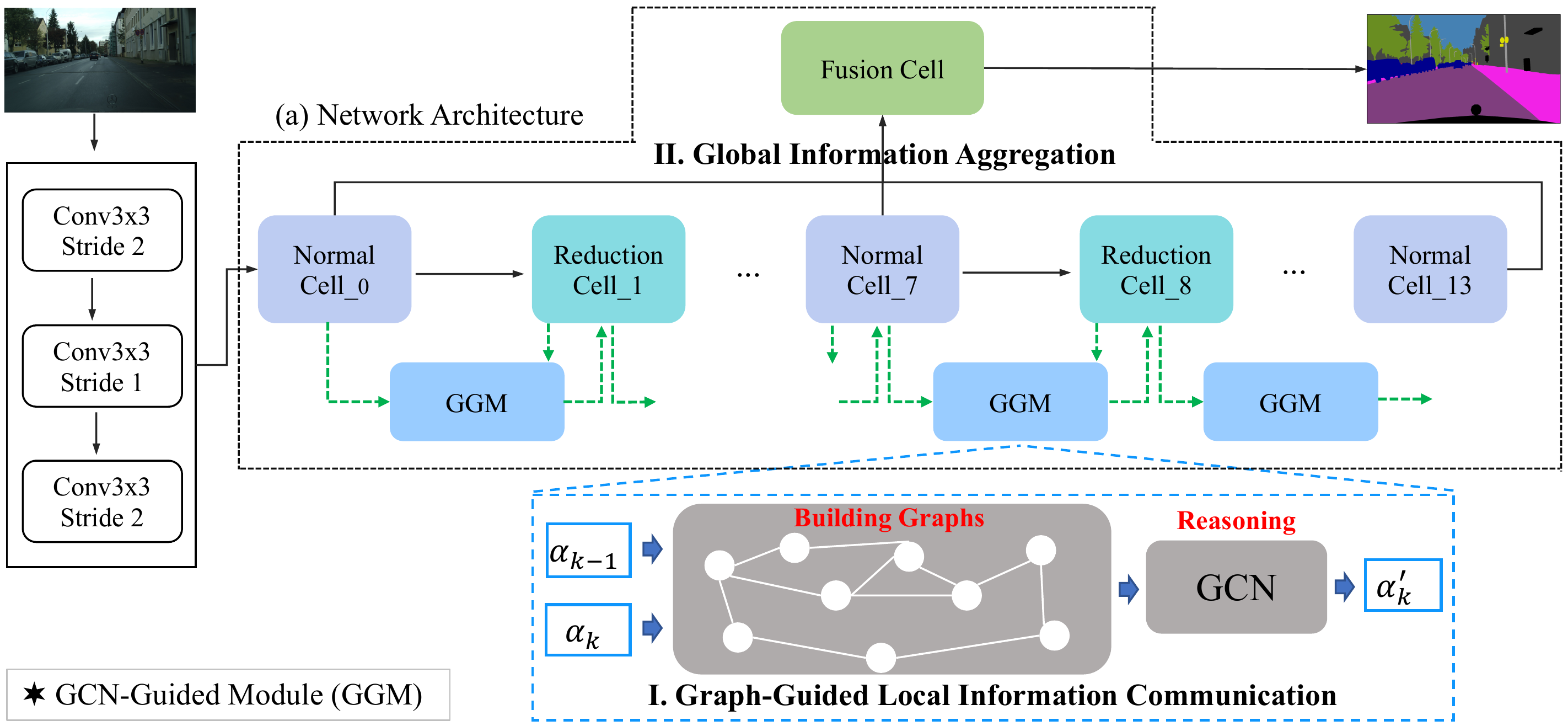}
  \end{center}
     \caption{Illustration of our local-to-global information communication network. In reduction cells, all the operations adjacent to the input nodes are of stride two. (a) The backbone network, which is stacked by a series of independent cells. (\uppercase\expandafter{\romannumeral1}) The GCN-Guided Module (GGM), which performs {\it \textbf{local}} information exchange between adjacent cells. $\alpha_{k}$ and $\alpha_{k-1}$ represent the architecture parameters for cell $k$ and cell $k-1$, respectively. $\alpha_{k}^{'}$ is the updated architecture parameters by GGM for cell $k$. The dotted lines indicate that GGM is only utilized in the search progress. (\uppercase\expandafter{\romannumeral2}) The Dense-Connected Fusion Cell, which aggregates the multi-scale feature to capture {\it \textbf{global}} context information.}
  \label{fig:ggn_network}
  \end{figure*}
  
  \section{Related Work}

  \paragraph{Semantic Segmentation Methods}
  FCN~\cite{long2015fully} is the pioneer work to achieve the end-to-end semantic segmentation task. Since then, to improve the segmentation performance, some remarkable works have utilized various heavy backbones \cite{corr_SimonyanZ14a,conf_cvpr_HeZRS16,journals_corr_HuangLW16a,DBLP:conf/cvpr/Chollet17} or well-designed network modules to capture multi-scale context information~\cite{zhao2017pyramid,DBLP:journals/corr/ChenPSA17,DBLP:conf/eccv/ChenZPSA18}. These outstanding works are designed for high-performance segmentation, which is inapplicable to the real-time applications. In terms of efficient segmentation methods, there are two mainstreams. One is to employ the human-designed and relatively light backbone (\eg~ENet~\cite{paszke2016enet}) or introduce some efficient operations (\eg~depth-wise dilated convolution). DFANet~\cite{li2019dfanet} utilizes a lightweight backbone to speed up and equips with a cross-level feature aggregation module to remedy the accuracy drop. Another is based on a multi-branch algorithm that consists of more than one path. For example, ICNet~\cite{zhao2018icnet} proposes to use the multi-scale image cascade to speed up the inference. BiSeNet~\cite{DBLP:conf/eccv/YuWPGYS18} decouples the extraction for spatial and context information using two paths.
 
  \paragraph{Information Communication for Semantic Segmentation}
  Capturing local details and global context information in images and properly combining them is critical for semantic segmentation performance. A post-process module of conditional random field (CRF) \cite{chen2017deeplab} is proposed to improve the ability of capturing local details. After that, the dilated convolution \cite{chen2018deeplab} is utilized to remedy the resolution loss, and the ASPP \cite{chen2018deeplab} and PSPNet \cite{zhao2017pyramid} modules enable efficient combination of local detail information and global context information based on the multi-scale features. 
  
  \paragraph{Neural Architecture Search} Neural Architecture Search (NAS) aims at automatically searching network architectures. Most existing architecture search works are based on either reinforcement learning~\cite{DBLP:conf/iclr/ZophL17,DBLP:journals/corr/abs-1812-05285} or evolutionary algorithm~\cite{DBLP:journals/corr/abs-1802-01548,DBLP:journals/corr/abs-1808-00193}. Though they can achieve satisfactory performance, they need thousands of GPU hours. To solve this time-consuming problem, one-shot methods \cite{bender2018understanding,brock2017smash} have been developed to train a parent network from which each sub-network can inherit the weights. They can be roughly divided into cell-based and layer-based methods according to the type of search space. For cell-based methods, ENAS~\cite{DBLP:conf/icml/PhamGZLD18} proposes a parameter sharing strategy among sub-networks, and DARTS~\cite{DBLP:conf/iclr/LiuSY19} relaxes the discrete architecture distribution as continuous deterministic weights, such that they could be optimized with gradient descent. SNAS~\cite{DBLP:conf/iclr/XieZLL19} proposes novel search gradients that train neural operation parameters and architecture distribution parameters in the same round of back-propagation. Additionally, there are some fantastic works~\cite{chen2019progressive,noy2019asap} that gradually lower the size of the search space in order to lessen the complexity of optimization. For the layer-based methods, FBNet~\cite{wu2019fbnet}, MnasNet~\cite{tan2019mnasnet}, ProxylessNAS~\cite{cai2018proxylessnas} use a multi-objective search approach that optimizes both accuracy and real-world latency. 
  
  \paragraph{NAS for Segmentation}~In the field of semantic segmentation, DPC~\cite{DBLP:conf/nips/ChenCZPZSAS18} is the pioneer work by introducing meta-learning techniques into the network architecture search problem. Auto-Deeplab~\cite{DBLP:journals/corr/abs-1901-02985} searches cell structures and the downsampling strategy together in the same round. More recently, CAS~\cite{zhang2019customizable} searches an architecture with customized resource constraint and a multi-scale module that has been widely used in the semantic segmentation field. \cite{nekrasov2019fast} over-parameterizes the architecture during the training via a set of auxiliary cells using reinforcement learning. 
  
  \paragraph{Graph Convolutional Network}~Convolutional neural networks on graph-structure data is an emerging topic in deep learning research.~Kipf \cite{kipf2016semi} presents a scalable approach for graph-structured data that is based on an efficient variant of convolutional neural networks which operate directly on graphs, for better information propagation.~After that, Graph Convolutional Networks (GCNs) \cite{kipf2016semi} is widely used in many domains, such as video classification \cite{wang2018videos} and action recognition \cite{stgcn2018aaai}. 
  Recently, Zhang et al. \cite{Zhang2019GraphHF} propose the Graph HyperNetwork to amortize the search cost: given an architecture, it directly generates the weights by running inference on a graph neural network. 
  In this paper, we apply the GCNs to model the relationship of adjacent cells in network architecture search. Specifically it takes the combination of local and global information into consideration during the search progress through the communication between adjacent cells and dense-connected fusion.


  \section{Methods}
  In this section, we first formulate the search problem and the overall framework, and then introduce the cell architecture search. Later we describe the proposed local to global information communication method, which consists of GCN-Guided Module (GGM) for local information exchange between adjacent cells, dense-connected fusion module for global information aggregation and latency-oriented search for a lightweight model, respectively.
  
  %
  %
  
  \subsection{Problem Formulation}
  
  \paragraph{Overview} As shown in Figure \ref{fig:ggn_network}(a), the input image is firstly processed by three convolutional layers followed by a series of independent cells, then the searchable dense-connected fusion cell fuses long-range multi-scale features for producing the representation feature by considering the local detail information and global context information before the pixel level classification. The GCN-Guided module is embedded into the search framework to bridge the information between adjacent cells. Then the search process is directed towards the goal of a lightweight network by the latency-oriented optimization loss.
  
  The overall loss function in the training stage can be formulated as:
  \begin{equation}\label{equ:loss}
  \mathop {\min }\limits_{a \in \mathcal{A}} L_{val} + \beta * L_{lat}
  \end{equation}
  where $\mathcal{A}$ denotes the search space, $L_{val}$ and $L_{lat}$ are the validation loss and the latency loss, respectively. Our goal is to search an optimal architecture ${a \in \mathcal{A}}$ that achieves the best trade-off between the performance and latency.

  \paragraph{Notation Table}
  
  As shown in Table \ref{tab:TableOfNotationForMyResearch}, we build the following notation table for clear representation.

  \begin{table}[h]\caption{The symbols and notations used in this paper.}
  \begin{center}
  \begin{tabular}{r c p{5cm} }
  \toprule
  ${N}$ & $\triangleq$ & the intermediate node number of the cell\\
  ${M}$ & $\triangleq$ & the number of candidate operations\\
  $\widetilde{O}_{h,i}$ & $\triangleq$ & the selected operation at edge ($h$, $i$) \\
  $Z_{h,i}$ & $\triangleq$ & the one-hot random variable at edge ($h$, $i$)\\
  ${O}_{h,i}$ & $\triangleq$ & all possible operations at edge ($h$, $i$)\\
  $\alpha_{h,i}$ & $\triangleq$ & the architecture parameter at edge ($h$, $i$)\\
  $\alpha_{k}$ & $\triangleq$ & the architecture parameter matrix of cell $k$\\
  $lat_{h,i}^{m}$ & $\triangleq$ & the latency cost of candidate operation $m$ at edge ($h$, $i$) \\
  
  \bottomrule
  \end{tabular}
  \end{center}
  \label{tab:TableOfNotationForMyResearch}
  \end{table}

  \subsection{Preliminaries: Cell Architecture Search}
  
  \begin{figure}[t]
  \centering
  \includegraphics[width=1.2in]{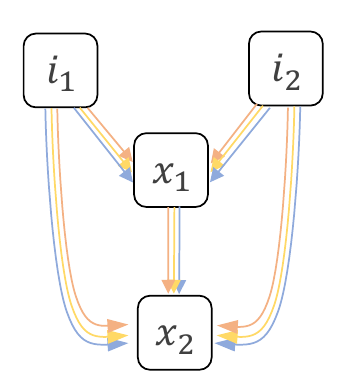}
  \caption{The structure of cell in our LGCNet. Each colored edge represents one candidate operation.}
  \label{fig:cell}
  \end{figure}
  
  A cell is the basic component of a network, which is a directed acyclic graph (DAG) as shown in Figure \ref{fig:cell}. Each cell has two input nodes $i_{1}$ and $i_{2}$, ${N}$ ordered intermediate nodes, denoted by $\mathcal{N} = \{x_1,...,x_N\}$, and an output node that outputs the concatenation of all intermediate nodes $\mathcal{N}$. Each node represents the latent representation (\eg~feature map) in the network, and each directed edge in this DAG represents a candidate operation (\eg~conv, pooling). The number of intermediate nodes ${N}$ is 2 in our work. Each intermediate node takes all its previous nodes as input. In this way, ${x_1}$ has two inputs $I_{1} = \{i_1, i_2\}$ and node ${x_2}$ takes $I_{2} = \{i_1, i_2, x_1\}$ as inputs. The intermediate nodes $x_{i}$ can be calculated by:
  
  \begin{equation}
  x_{i}  \! =  \! \sum_{c \in I_{i}} \widetilde{O}_{h,i}(c)
  \end{equation}
  where $\widetilde{O}_{h,i}$ is the selected operation at edge ($h$, $i$). 
  
  To search the selected operation $\widetilde{O}_{h,i}$, the search space is represented with a set of one-hot random variables from a fully factorizable joint distribution $p(Z)$ \cite{DBLP:conf/iclr/XieZLL19} and is optimized by the single level optimization as opposed to bilevel optimization employed by DARTS \cite{DBLP:conf/iclr/LiuSY19}. Concretely, each edge is associated with a one-hot random variable which is multiplied as a mask to the all possible operations ${O}_{h,i}$ = ($o^{1}_{h,i}$, $o^{2}_{h,i}$, ..., $o^{M}_{h,i}$) in this edge. We denote the one-hot random variable as $Z_{h,i}$ = ($z^{1}_{h,i}$, $z^{2}_{h,i}$, ..., $z^{M}_{h,i}$) where ${M}$ is the number of candidate operations. The intermediate nodes during search process in such way are:
  
  \begin{small}
  \begin{equation}\label{intracell}
  x_{i}  \! =  \! \sum_{c \in I_{i}} \widetilde{O}_{h,i}(c) = \sum_{c \in I_{i}} \sum_{m=1}^{M} z_{h,i}^m o_{h,i}^m(c)
  \end{equation}
  \end{small}
  
  An essential issue is how to make the one-hot random variables $z$ (or $P(Z)$) differentiable in Equation \ref{intracell}. We use reparameterization \cite{maddison2016concrete} to relax the discrete architecture distribution to be continuous:
  \begin{small}
  \begin{equation}\label{equ:softmax}
  Z_{h,i} \! = f_{\alpha_{h,i}}(G_{h,i}) = \text{Softmax}((log \alpha_{h,i} + G_{h,i}) / \lambda)
  \end{equation}
  \end{small}
  where $\alpha_{h,i}$ is the architecture parameters at the edge $(h,i)$, and $G_{h,i}$ = $-log(-log(U_{h,i}))$ is a vector of Gumbel random variables, $U_{h,i}$ is a uniform random variable and $\lambda$ is the temperature of softmax.
  
  To better balance the speed and performance, we only employ the following 8 types of operations in Table~\ref{Section 3.3} as the set of candidate operations $O$:
  
   \begin{table}[h]\caption{The candidate operations for the cell-independent search.}
  \begin{center}
  \begin{tabular}{l}
  \toprule
  3 $\times$ 3 max pooling \\
  skip connection \\
  3 $\times$ 3 conv \\
  zero operation \\
  3 $\times$ 3 separable conv \\
  3 $\times$ 3 dilated separable conv (dilation=2) \\
  3 $\times$ 3 dilated separable conv (dilation=4) \\
  3 $\times$ 3 dilated separable conv (dilation=8) \\
  \bottomrule
  \end{tabular}
  \end{center}
  \label{Section 3.3}
  \end{table}
  
  
  \subsection{Local to Global Information Communication}
  
  \paragraph{GCN-Guided Module for Local Information Exchange}
  \par With cell independent mechanism, we propose a novel GCN-Guided Module (GGM) to naturally bridge the operation information between adjacent cells. The total network architecture of our GGM is shown in Figure \ref{fig:ggn_network} (\uppercase\expandafter{\romannumeral1}). Inspired by \cite{wang2018videos}, the GGM represents the communication between adjacent cells as a graph and performs reasoning on the graph for information exchange. For more through explanation on what role that GGM plays, please refer to the section $D$ of Network Visualization and Analysis.
  
  \par Specifically, we utilize the similarity relations of edges in adjacent cells to construct the graph where each node represents one edge in cells. In this way, the state changes for the previous cell can be delivered to the current cell by reasoning on this graph. As stated before, let $\alpha_{k}$ represents the architecture parameter matrix for the cell $k$, and the dimension of $\alpha_{k}$ is $p$ $\times$ $q$ where $p$ and $q$ represents the number of edges and the number of candidate operations, respectively. Same as cell $k$, the architecture parameter $\alpha_{k-1}$ for cell $k-1$ is also a $p$ $\times$ $q$ matrix. To fuse the architecture parameter information of previous cell $k-1$ into the current cell $k$ and generate the updated $\alpha'_{k}$, we model the information propagation between cell ${k-1}$ and cell $k$ as follows:
  
  \begin{small}
  \begin{equation}\label{equ:residual1}
  \alpha'_{k} =  \alpha_{k}  +  \gamma \Phi_{2}(G(\Phi_{1}(\alpha_{k-1}), \mathbf{A})) \\
  \end{equation}
  \end{small}
  where $\mathbf{A}$ represents the adjacency matrix of the reasoning graph between cells $k$ and $k-1$, and the function $G$ denotes the Graph Convolutional Networks (GCNs)~\cite{kipf2016semi} to perform reasoning on the graph. $\Phi_{1}$ and $\Phi_{2}$ are two different transformations with two fully connected (FC) layers. Specifically, $\Phi_{1}$ maps the original architecture parameter to the embedding space and $\Phi_{2}$ transfers it back into the source space after the GCN reasoning. $\gamma$ controls the fusion of two types of architecture parameter information.
  
  For the function $G$, we construct the reasoning graph between cell ${k-1}$ and cell $k$ by their similarity. Given an edge in cell $k$, we calculate the similarity between this edge and all other edges in cell $k-1$, and a softmax function is used for normalization. Therefore, the adjacency matrix $\mathbf{A}$ of the graph between two adjacent cells $k$ and $k-1$ can be established by:
  
  \begin{equation}\label{equ:adject}
    \mathbf{A} = \text{Softmax}( \phi_1 (\alpha_{k}) * \phi_2 ({ \alpha_{k-1}}) ^{T} )
  \end{equation}
  where we have two different transformations $\phi_1$ = $\alpha_{k}{w_1}$ and $\phi_2$ = $\alpha_{k-1}{w_2}$ for the architecture parameters, and parameters $w_1$ and $w_2$ are both $q \times q$ weights which can be learned via back propagation. The result $\mathbf{A}$ is a $p \times p$ matrix.
  
  Based on this adjacency matrix $\mathbf{A}$, we use the GCNs to perform information propagation on the graph as shown in Equation \ref{equ:gcn}. A residual connection is added to each layer of GCNs. The GCNs allow us to compute the response of a node based on its neighbors defined by the graph relations, so performing graph convolution is equivalent to performing message propagation on the graph.
  
  \begin{small}
  \begin{equation}\label{equ:gcn}
  G(\Phi_{1}(\alpha_{k-1}), \mathbf{A}) = \mathbf{A} \Phi_{1}(\alpha_{k-1}) W_{k-1}^{g}  + \Phi_{1}(\alpha_{k-1})
  \end{equation}
  \end{small}
  where the $W_{k-1}^{g}$ denotes the GCNs weight with dimension $d \times d$. In fact, the weight $W$ in GCN is embedded into the updated architecture parameter (\ie $\alpha'$), as shown in Equation \ref{equ:residual1}, thus the weight $W$ can be learned via back propagation.
  
  The proposed GGM seamlessly integrates the graph convolutional network into neural architecture search, which can bridge the operation information between adjacent cells. Moreover, the GCN-guided Module is only used in the training phase, thus it introduces no extra parameters and computational cost to the searched model in the inference stage.

  \paragraph{Dense-Connected Fusion Cell for Global Information Aggregation}
  
  \begin{figure}[t]
  \centering
  \includegraphics[width=1.0\columnwidth]{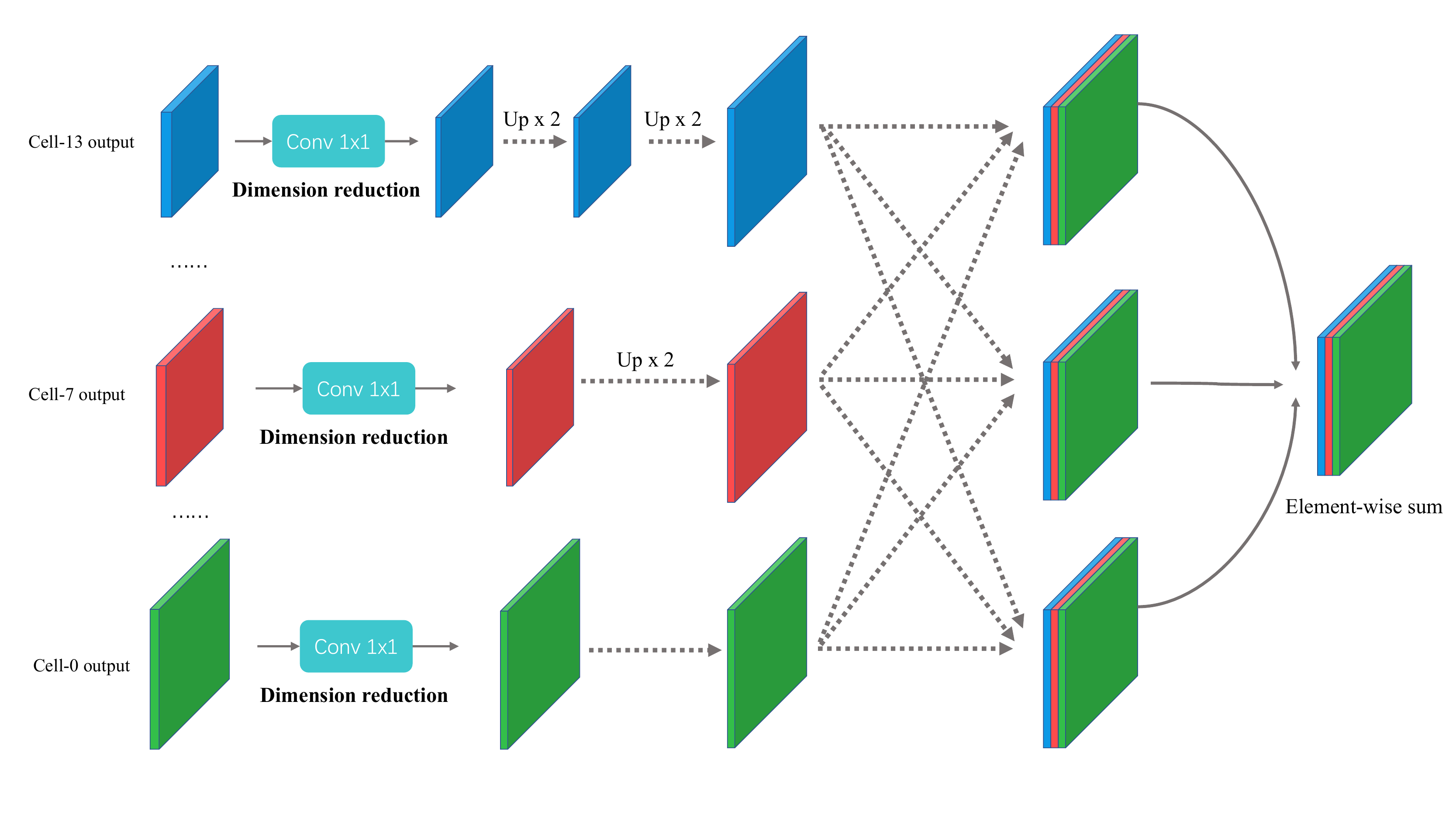}
  \caption{Overview of the dense-connected fusion cell for automatic multi-scale feature aggregation. The fusion cell contains E edges (dotted arrows), each edge is equipped with some candidate operations. ``Up$\times$2'' means upsampling operation.}
  \label{fig:fusioncell}
  \end{figure}
  
  
  As we mentioned before, the effective combination of local details and global context information is crucial for semantic segmentation, which is usually accomplished by the multi-scale module such as ASPP in Deeplab and multi-scale cell in CAS. However, multi-scale cell in CAS only considers short range features from stride=8 and stride=16 to reduce time cost, while the ASPP also takes more latency due to multiple convolutional layers. We thus propose the dense-connected fusion cell to tackle these two issues. On one hand, our dense-connection cell aggregate long range features from stride=4 to stride=16 in the network, which can capture more fine-grained clues to further improve accuracy. On the other hand, we search operations for each feature scales, which can obtain more lightweight operations guided by the latency oriented loss (convolutional layer will obtain larger loss than lightweight operations such as pooling).
  

  The structure of the proposed dense-connected fusion cell is shown in Figure \ref{fig:fusioncell}. As shown in Figure \ref{fig:ggn_network}, the fusion cell takes the outputs of cell-0, cell-7, and cell-13 with different resolutions as its inputs, thus the dense-connected fusion cell is designed to combine multi-scale features (\ie, low-level spatial details and high-level semantic context). The dense-connected fusion cell is designed as a directed acyclic graph consisting of M nodes and E edges, each node is a latent representation (\ie, feature map) and each directed edge is associated with some candidate operations. As shown in Figure \ref{fig:fusioncell}, the number of channels of three inputs will be reduced to 48 through a $1$ $\times$ $1$ convolution layer to save time cost in subsequent process. Each edge is able to search specific operators from the search space, unless explicitly specified by “Up$\times$2”, in which only deconvolutional layer and upsample layer (interpolate) can be searched. The middle feature maps of three branches are cross-branch connected densely by concatenation. Then the output of the dense-connected fusion cell is designed as the element-wise sum of the final feature maps from three branches. We use the same sampling and optimization process as the normal/reduction cell to optimize the fusion cell’s architecture parameter.
  
  Given the candidate operation set, the dense-connected fusion cell also efficiently enlarges the receptive field of the network. For the operation set of the aggregation cell, we collect the following 10 kinds of operations in Table~\ref{Section 3.4}:
  
  \begin{table}[h]\caption{The candidate operations for the aggregation cell.}
  \begin{center}
  \begin{tabular}{l}
  \toprule
  1 $\times$ 1 conv \\
  3 $\times$ 3 conv \\
  3 $\times$ 3 separable conv \\
  3 $\times$ 3 dilated separable conv (dilation=2) \\
  3 $\times$ 3 dilated separable conv (dilation=4) \\
  3 $\times$ 3 dilated separable conv (dilation=8) \\
  3 $\times$ 3 dilated separable conv (dilation=12) \\
  Global pooling with output size 1 $\times$ 1 + 1 $\times$ 1 conv + upsampling  \\
  Global pooling with output size 2 $\times$ 2 + 1 $\times$ 1 conv + upsampling \\
  Global pooling with output size 5 $\times$ 5 + 1 $\times$ 1 conv + upsampling \\
  \bottomrule
  \end{tabular}
  \end{center}
  \label{Section 3.4}
  \end{table}
  
  
  \paragraph{Latency-Oriented Search}
  
  To obtain a real-time semantic segmentation network, we take the real-world latency into consideration during the search process, which guides the search process towards the direction to find an optimal lightweight model. Specifically, we create a GPU-latency lookup table \cite{cai2018proxylessnas,wu2019fbnet,zhang2019customizable,tan2019mnasnet} which records the inference latency of each candidate operation. During the search process, each candidate operation $m$ at edge ($h$, $i$) will be assigned a cost $lat_{h,i}^{m}$ given by the pre-built lookup table. In this way, the total latency for cell $k$ is accumulated as:
  
  \begin{equation}\label{equ:latency}
  lat_k \! = \! \sum_{h,i}\sum_{m=1}^{M} z_{h,i}^{m} lat_{h,i}^{m}
  \end{equation}
  where $z_{h,i}^{m}$ is the softened one-hot random variable as stated in Equation \ref{intracell}. Given an architecture $a$, the total latency cost is estimated as:
  
  \begin{equation}\label{equ:latencym}
  LAT(a) \! = \! \sum_{k=0}^{K} lat_{k}
  \end{equation}
  where $K$ refers to the number of cells in architecture $a$. The latency for each operation $lat_{h,i}^{m}$ is a constant and thus total latency loss is differentiable with respect to the architecture parameter $\alpha_{h,i}$.
  
  The total loss function in LGCNet is designed as follows:
  \begin{equation}
  \label{equ:latency2}
      \begin{split}
      L(a, w) &= \min \limits_{a \in \mathcal{A}} L_{val} + \beta * L_{lat} \\
      &= \min \limits_{a \in \mathcal{A}} CE(a, w_a) + \beta ~log(LAT(a))
      \end{split}
  \end{equation}
  where $CE(a, w_a)$ denotes the cross-entropy loss of architecture $a$ with parameter $w_a$, $LAT(a)$ denotes the overall latency of architecture $a$, which is measured in micro-second, and the coefficient $\beta$ controls the balance between the accuracy and latency. The architecture parameter $\alpha$ and the weight $w$ are optimized in the same round of back-propagation.

  \section{Experiments}
   In this section, we conduct extensive experiments to verify the effectiveness of our LGCNet and compare the network searched by our method with other works on two standard benchmarks. 
   
  \subsection{Experiment Setup}
  
  \paragraph{Datasets and Evaluation Metrics} In order to verify the effectiveness and robustness of our method, we evaluate it on the Cityscapes \cite{cordts2016cityscapes} and CamVid~\cite{Camvid} datasets. Cityscapes \cite{cordts2016cityscapes} is a public released dataset for urban scene understanding. It contains 5000 high-quality pixel-level fine annotated images (2975, 500, and 1525 for the training, validation, and test sets, respectively) with size 1024 $\times$ 2048 collected from 50 cities. The dense annotation contains 30 common classes and 19 of them are used in training and testing. CamVid~\cite{Camvid} is another public released dataset with object class semantic labels. It contains 701 images in total, in which 367 for training, 101 for validation and 233 for testing. The images have a resolution of 960 $\times$ 720 and 11 semantic categories. For evaluation, we use the mean of class-wise intersection over union (mIoU), network forward time (Latency), and Frames Per Second (FPS) as the evaluation metrics.

  \paragraph{Implementation Details}
  
  \par We conduct all the experiments using Pytorch 0.4 \cite{pytorch} on a workstation, and the inference time for all the experiments is reported on one Nvidia Titan Xp GPU. The whole pipeline contains three sequential steps: search, pretraining and finetuning. It starts with the search process on the target dataset and obtains the light-weight architecture according to the optimized $\alpha$ followed by the ImageNet~\cite{deng2009imagenet} pretraining, and this pretrained model is subsequently finetuned on the specific dataset for 200 epochs.
  
  In the search process, the architecture contains 14 cells and each cell has $N$ = 2 intermediate nodes. With the consideration of speed and accuracy, we set the initial channel for the network as 8. For the training hyper-parameters, the mini-batch size is set to 16. The architecture parameters $\alpha$ are optimized by Adam, with initial learning rate 0.001, $\beta$ = (0.5, 0.999) and weight decay 0.0001. The network parameters are optimized using SGD with momentum 0.9, weight decay 0.001, and cosine learning scheduler that decays learning rate from 0.025 to 0.001. For gumbel softmax, we set the initial temperature $\lambda$ in equation \ref{equ:softmax} as 1.0, and it gradually decreases to the minimum value of 0.03. The search time cost on Cityscapes takes approximately 10 hours with 16 TitanXP GPU cards.
  
  For finetuning details, we train the network with mini-batch 8 and SGD optimizer with `poly' scheduler that the learning rate decays from 0.01 to zero. Following \cite{DBLP:journals/corr/WuSH16a}, the online bootstrapping strategy is applied to the finetuning process. For data augmentation, we use random flip and random resize with a scale between 0.5 and 2.0. Finally, we randomly crop the image with a fixed size for training.
  
  For the GCN-guided Module, we use one Graph Convolutional Network (GCN)~\cite{kipf2016semi} between two adjacent cells, and each GCN contains one layer of graph convolutions. The kernel size of the GCN parameters $W$ in equation \ref{equ:gcn} is 64 $\times$ 64. We set the $\gamma$ as 0.5 in equation \ref{equ:residual1} in our experiments.
  
  For the dense-connected fusion cell, we set the node number $M$ and edge number $E$ as 9 and 13, respectively.
  
  \begin{table}[h]
  \caption{Comparing results on the Cityscapes test set. Methods trained using both fine and coarse data are marked with $*$. The mark $\S$ represents that the speed is measured on TitanX, and the mark ${\dag}$ denotes the speed is remeasured on Titan Xp. The mark $\tau$ indicates that TensorRT acceleration has been used.}
  \begin{center}
  \setlength{\tabcolsep}{1mm}{
  \begin{tabular}{|l|c|c|c|c|}
  \hline
  Method  & Input Size & mIoU (\%) & Latency(ms) & FPS \\
  \hline\hline
  FCN-8S \cite{long2015fully} & 512 $\times$ 1024  & 65.3 & 227.23 & 4.4    \\
  PSPNet \cite{zhao2017pyramid}  & 713 $\times$ 713  & 78.4 & 1288.0 & 0.78   \\
  DeepLabV3$^*$ \cite{DBLP:journals/corr/ChenPSA17} & 769 $\times$ 769   & 81.3 & 769.2 & 1.3  \\
  AutoDeepLab$^*$ $^{\dag}$ \cite{DBLP:journals/corr/abs-1901-02985} & 769 $\times$ 769 & 81.2 & 303.0 & 3.3  \\
  \hline
  SegNet  \cite{badrinarayanan2017segnet}      & 640 $\times$ 360   & 57.0 & 30.3   & 33     \\
  ENet  \cite{paszke2016enet}         & 640 $\times$ 360   & 58.3 & 12.7   & 78.4   \\
  SQ  \cite{treml2016speedingSQ}          & 1024 $\times$ 2048 & 59.8 & 46.0   & 21.7   \\
  ICNet  \cite{zhao2018icnet}       & 1024 $\times$ 2048 & 69.5 & 26.5   & 37.7   \\
  SwiftNet \cite{SwiftNet} & 1024 $\times$ 2048 & 75.1 & 26.2   & 38.1 \\
  DF1-Seg \cite{li2019partialdfnet} & 768 $\times$ 1536 & 73.0 & 29.1 & 34.4   \\
  ESPNet \cite{mehta2018espnet}  & 1024 $\times$ 512  & 60.3 & 8.2    & 121.7 \\
  ESPNetV2 \cite{Mehta2019ESPNetv2AL}  & 1024 $\times$ 512  & 66.2 & 14.7 & 68.0 \\
  ERFNet \cite{Romera2018ERFNetER}  & 1024 $\times$ 512  & 69.7 & 48.5   & 20.6 \\
  BiSeNet-Xception39  \cite{DBLP:conf/eccv/YuWPGYS18}  & 768 $\times$ 1536  & 68.4 & 9.5 & 105.8  \\
  BiSeNet-Res18  \cite{DBLP:conf/eccv/YuWPGYS18}  & 768 $\times$ 1536  & 74.7 & 15.3   & 65.5  \\
  DFANet A$\S$  \cite{li2019dfanet}   & 1024 $\times$ 1024 & 71.3 & 10.0  & 100.0   \\
  DFANet A $^{\dag}$ \cite{li2019dfanet} \footnote[1] & 1024 $\times$ 1024 & 71.3 & 19.0  & 52.6   \\
  ShelfNet \cite{Zhuang2019ShelfNetFF} & 768 $\times$ 1536 & 74.8 & 16.9 & 59.2   \\
  CAS \cite{zhang2019customizable} & 768 $\times$ 1536  & 70.5 & 9.25   & 108.0  \\
  CAS$^*$ \cite{zhang2019customizable} & 768 $\times$ 1536  & 72.3 & 9.25   & 108.0  \\
  GAS \cite{gas} & 769 $\times$ 1537  & 71.8 & 9.22 & 108.4  \\
  GAS$^*$ \cite{gas} & 769 $\times$ 1537  & 73.5 & 9.22   & 108.4  \\
  FasterSeg \cite{Chen2020FasterSegSF} \footnote[2] & 1024  $\times$ 2048 & 71.5 & 11.13 & 89.8  \\
  HyperSeg-M \cite{nirkin2021hyperseg} & 512 $ \times$ 1024 & 75.8 & 21.6 & 46.2  \\
  BiSenet v2$^\tau$ \cite{yu2021bisenet} & 512 $ \times$ 1024 & 72.6 & 5.6 & 179.2  \\
  STDC1-Seg75$^\tau$ \cite{fan2021rethinking} & 768 $ \times$ 1536 & 75.3 & 6.98 & 143.1  \\
  \hline
  \textbf{LGCNet}  & 769 $\times$ 1537  & 72.4 & 8.68  & 115.2  \\
  \textbf{LGCNet$^*$} & 769 $\times$ 1537  & 74.0 & 8.68 & 115.2  \\
  \textbf{LGCNet$^\tau$} & 769 $\times$ 1537  & 74.0 & 4.67 & 205.6  \\
  \hline
  \end{tabular}}
  \end{center}
  \label{Cityscapes}
  \end{table}

  \subsection{Real-time Semantic Segmentation Results}
  
  In this part, we compare the model searched by LGCNet with other existing real-time segmentation methods on semantic segmentation datasets. The inference time is measured on an Nvidia Titan Xp GPU and the speed of other methods reported on Titan Xp GPU in CAS \cite{zhang2019customizable} are used for fair comparison. Moreover, the speed is remeasured on Titan Xp if the origin paper reports it on different GPU and is not mentioned in CAS \cite{zhang2019customizable}.
  
  \paragraph{Results on Cityscapes} We evaluate the network searched by LGCNet on the Cityscapes test set. The validation set is added to train the network before submitting it to Cityscapes online server. Following \cite{DBLP:conf/eccv/YuWPGYS18,zhang2019customizable}, LGCNet takes an input image with size 769 $\times$ 1537 that is resized from origin size 1024 $\times$ 2048. Overall, our LGCNet gets the best performance among all methods with the speed of 115.2 FPS. With only fine data and without any evaluation tricks, our LGCNet yields 72.4\% mIoU which is the state-of-the-art trade-off between performance and speed for real-time semantic segmentation. 
LGCNet achieves 74.0\% when the coarse data is utilized. The full comparison results are shown in Table \ref{Cityscapes}. Compared to BiSeNet-Xception39 \cite{DBLP:conf/eccv/YuWPGYS18} and CAS \cite{zhang2019customizable} that have comparable running speed with us, our LGCNet surpasses them by a large margin with 4.0\% and 1.9\% improvement, respectively. Compared to other methods such as SegNet \cite{badrinarayanan2017segnet}, ENet \cite{paszke2016enet}, SQ \cite{treml2016speedingSQ} and ICNet \cite{zhao2018icnet}, our method achieves significant improvement in speed while achieving the performance gain over them about 15.4\%, 14.1\%, 12.6\%, 2.9\%, respectively. To fairly compare with BiSeNet v2 \cite{yu2021bisenet} and STDC1-Seg75 \cite{yu2021bisenet}, we remeasured our method with TensorRT $5.1.5$ acceleration. Although STDC1-Seg75 \cite{yu2021bisenet} performs better than our method in accuracy, our running speed is much faster (205.6 vs 143.1 in FPS), which demonstrates that the proposed method can achieve either better performance or faster running speed when comparing with the co-occurrent works.
  
  \footnotetext[1]{After merging the BN layers for DFANet, there still has a speed gap between the original paper and our measurement. We suspect that it is caused by the inconsistency of the implementation platform in which DFANet has the optimized depth-wise convolution (DW-Conv). LGCNet} also have many candidate operations using DW-Conv, so the speed of our LGCNet is still capable of beating it if the DW-Conv is optimized correctly like DFANet.

  \footnotetext[2]{The speed, which is tested by Pytorch without TensorRT, is provided by the authors.}
  
  \paragraph{Results on CamVid} To evaluate the transfer capacity of LGCNet, we transfer the network searched on Cityscapes to Camvid directly. Table \ref{CamVid} shows the comparison results with other methods. With input size 720 $\times$ 960, LGCNet achieves the 73.8\% mIoU with 158.6 FPS that is also the state-of-the-art trade-off between performance and speed, which demonstrates the superior transferability of LGCNet.

  \begin{table}[h]
  \caption{Results on the CamVid test set with resolution of 960 $\times$ 720. "-" indicates the corresponding result is not provided by the methods.}
  \begin{center}
  \scalebox{1.1}{
  \setlength{\tabcolsep}{1mm}{
  \begin{tabular}{|l|c|c|c|}
  \hline
  Method  & ~mIoU (\%)~ & Latency(ms) & ~FPS~ \\
  \hline\hline
  SegNet \cite{badrinarayanan2017segnet}   & 55.6 & 34.01 & 29.4   \\
  ENet \cite{paszke2016enet}  & 51.3 & 16.33 & 61.2   \\ 
  ICNet \cite{zhao2018icnet} & 67.1 & 28.98 & 34.5   \\
  BiSeNet  \cite{DBLP:conf/eccv/YuWPGYS18} & 65.6 &  -    & -   \\
  DFANet A \cite{li2019dfanet} & 64.7 & 8.33  & 120    \\
  CAS \cite{zhang2019customizable} & 71.2 & 5.92 & 169    \\
  FasterSeg \cite{Chen2020FasterSegSF} & 71.1 & 7.59 & 131.8  \\
  GAS \cite{gas} & 72.8 & 6.53 & 153.1  \\
  \hline
  LGCNet   & 73.8 & 6.31  & 158.6  \\
  \hline
  \end{tabular}}
  }
  \end{center}
  \label{CamVid}
  \end{table}

\begin{figure}[h]
  \centering
  \includegraphics[width=3.5in]{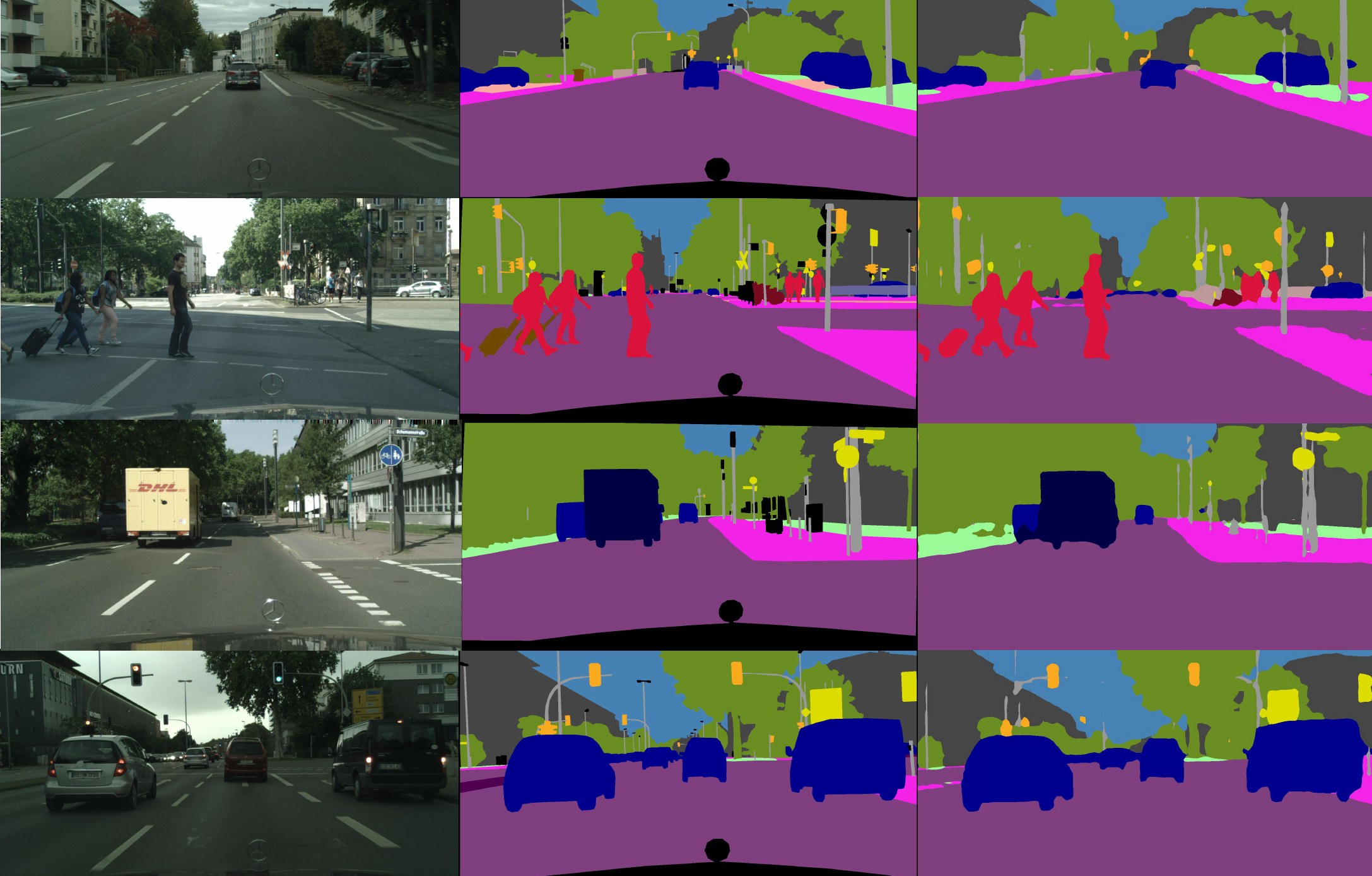}
  \caption{Results of the proposed LGCNet on the Cityscapes validation set. The first column is input image, the second column is ground truth, and the final column displays the output of LGCNet.}
  \label{fig:show_citys}
  \end{figure}

  \paragraph{Visual Segmentation Results} We provide some visual prediction results on the Cityscapes validation set. As shown in Figure \ref{fig:show_citys}, the columns correspond to the input image, ground truth, the prediction of LGCNet. It demonstrates that LGCNet can achieve satisfactory and consistent visual results.
  
  \subsection{Ablation Study}
  
  \par To verify the effectiveness of each component in our framework, extensive ablation studies for the GCN-Guided module, dense-connected fusion cell, and the latency loss are performed. In addition, we also provide some insights about the role of the GCN-Guided Module in the search process.
  
  \paragraph{Effectiveness of GCN-Guided Module}
  \label{aba-GGM}
  We propose the GCN-Guided Module (GGM) to build the connection between cells. To verify the effectiveness of the GGM, we conduct a series of experiments with different strategies: a) network stacked by shared cells, \ie, one type of normal cell and one type of reduction cell; b) network stacked by independent cells, \ie, each cell has its own structure; c) based on strategy-b, using fully connected layer to infer the relationship between cells; d) based on strategy-b, using GGM to infer the relationship between cells. The corresponding experiment results are shown in Table \ref{tab:effi_gcn}. The performance reported here is the average mIoU over five repeated experiments on the Cityscapes validation set with latency loss weight $\beta$ = 0.005. 
$\pm$ denotes the variance of mIoU for each strategy. 
Overall, performance suffers greatly when there is just one independent cell since there is a larger search space, which makes optimization much more challenging. This performance drop is mitigated by adding a communication mechanism between cells. Especially, our GCN-Guided Module can bring about 3\% performance improvement compared to the fully-connected mechanism (\ie~setting (c)).

  \begin{table}[h]
    \caption{Ablation study for the effectiveness of GCN-Guided Module on Cityscapes validation dataset.}
    \begin{center}
    \setlength{\tabcolsep}{1mm}{
    \scalebox{1.1}{
    \begin{tabular}{|l|c|c|c|}
    \hline
    Methods  & mIoU (\%) & Params & FPS \\
    \hline\hline
   a) cell shared & 68.6 ($\pm$ 0.9) & 6.60 M & 100.1 \\
   b) cell independent  & 67.1 ($\pm$ 1.0) & 4.18 M & 118.9\\
   c) cell independent + FC  & 69.7 ($\pm$ 0.6) & 3.24 M & 112.8 \\
   d) cell independent + GCN  & \textbf{73.0} ($\pm$ 0.5) & 2.09 M & 115.2\\
    \hline
    \end{tabular}}
    }
    \end{center}
    \label{tab:effi_gcn}
    \end{table}


  \paragraph{Comparison against Random Search}
  
  As discussed in \cite{DBLP:conf/uai/LiT19}, random search is a competitive baseline for hyper-parameter optimization. To further verify the effectiveness of the GCN-Guided Module, we randomly sample ten architectures from the search space and evaluate them on the Cityscapes validation set with ImageNet pretrain. Specifically, we try two types of random settings in our experiments: a) fully random search without any constraint; b) randomly select the networks that meet the speed requirement over 100 FPS from the search space. The results are shown in Table \ref{randomsearch}, in which each value is the average result of ten random architectures. In summary, the network searched by LGCNet can achieve an excellent trade-off between performance and latency, while random search will result in high overhead without any latency constraint or low performance with latency constraint.
  
  \begin{table}[h]
  \caption{Comparison to random search on the Cityscapes validation set.}
  \begin{center}
  \setlength{\tabcolsep}{1mm}{
  \scalebox{1.1}{
  \begin{tabular}{|l|c|c|}
  \hline
  Methods  & mIoU (\%) & FPS \\
  \hline\hline
  Random setting (a)  & 69.8 ($\pm$ 0.5) & 61.9 ($\pm$ 4.8) \\
  Random setting (b)  & 66.3 ($\pm$ 0.4) & 107.6 ($\pm$ 3.4) \\
  LGCNet  & \textbf{73.0} ($\pm$ 0.5)  & 115.2 ($\pm$ 3.7) \\
  \hline
  \end{tabular}}
  }
  \end{center}
  \label{randomsearch}
  \end{table}
  
  \paragraph{Dimension Selection}
  
  The dimension selection of GCN weight $W$ in Equation \ref{equ:gcn} is also important, thus we conduct experiments with different GCN weight dimensions (denoted by $d$). Experimental results are shown in Table \ref{GCN_channel} in which the values are the average mIoU over five repeated experiments on the Cityscapes validation set with latency loss weight $\beta$ = 0.005. The experimental result indicates that LGCNet achieves the best performance when d = 64.
  
  \begin{table}[h]
  \caption{Ablation study for different GCN weight dimensions of GGM.}
  \begin{center}
  \setlength{\tabcolsep}{1mm}{
  \scalebox{1.1}{
  \begin{tabular}{|l|c|c|}
  \hline
  Methods  & mIoU (\%) & FPS \\
  \hline\hline
  without GGM & 67.1 & 112.9\\
  \hline
  GGM with d = 16  & 72.1  & 111.6      \\
  GGM with d = 32  & 72.2  & 109.2      \\
  GGM with d = 64  & \textbf{73.0}  & 115.2      \\
  GGM with d = 128 & 72.5  & 107.1      \\
  GGM with d = 256 & 72.6  & 116.3      \\
  \hline
  \end{tabular}}
  }
  \end{center}
  \label{GCN_channel}
  \end{table}
  
  \paragraph{Different Reasoning Graph}
  
  For GCN-Guided Module, in addition to the way described in Section 3.2, we also try another way to construct the reasoning graph. Specifically, we treat each candidate operation in a cell as a node in the reasoning graph. Given the architecture parameter $\alpha_{k}$ for cell $k$ with dimension $p \times q$, we first flatten the $\alpha_{k}$ and $\alpha_{k-1}$ to the one dimensional vector $\alpha'_{k}$ and $\alpha'_{k-1}$, and then perform matrix multiplication to get adjacent matrix $\mathbf{A} = \alpha'_{k} (\alpha'_{k-1})^{T}$. Different from the ``edge-similarity'' reasoning graph in Section 3.2, we call this graph ``operation-identity'' reasoning graph. We conduct the comparison experiment for two types of graphs on the Cityscapes validation set under the same latency loss weight $\beta$ = 0.005, the comparison results are shown in Table \ref{reasongraph}.
  
  \begin{table}[h]
  \caption{The comparison results for reasoning graph for edges and operations.}
  \begin{center}
  \setlength{\tabcolsep}{1mm}{
  \scalebox{1.1}{
  \begin{tabular}{|l|c|c|c|}
  \hline
  Reasoning Graph  &  mIoU (\%)  & FPS \\
  \hline\hline
  Operation-identity & 71.2 & 106.4   \\
  Edge-similarity    & \textbf{73.0} & 115.2   \\
  \hline
  \end{tabular}}
  }
  \end{center}
  \label{reasongraph}
  \end{table}
  
  Intuitively, the ``operation-identity'' way provides more fine-grained information about operation selection for other cells, while it also breaks the overall properties of an edge, and thus doesn't consider the other operation information at the same edge when making a decision. After visualizing the network, we also found that the ``operation-identity'' reasoning graph tends to make cell select the same operation for all edge, which increases the difficulty of the trade-off between performance and latency. This can also be verified from the result in Table \ref{reasongraph}. So we choose the ``edge-similarity'' way to construct the reasoning graph as described in Section 3.2.
  
  \paragraph{Effectiveness of the Dense-connected Fusion Cell}
  
  To demonstrate the effectiveness of the proposed dense-connected fusion cell (DCFC), we conduct a series of experiments with different strategies: a) without multi-scale feature aggregation; b) with ASPP module \cite{DBLP:journals/corr/ChenPSA17}; c) with PPM module \cite{zhao2017pyramid}; d) with searched MSCell in \cite{zhang2019customizable}; e) with searched dense-connected fusion cell (DCFC). The results are shown in Table \ref{ppm}. Overall, the searched dense-connected fusion cell successfully boosts up the mIoU from 69.5\% to 73.0\% on the Cityscapes validation set. Particularly, the searched aggregation cell surpasses the ASPP module and PPM module, searched MSCell by 0.6\% and 0.5\%, 0.3\% performance gains with faster inference speed. 
  
  \begin{table}[h]
  \caption{The performance for different multi-scale modules on the Cityscapes validation set.}
  \begin{center}
  \setlength{\tabcolsep}{1mm}{
  \scalebox{1.1}{
  \begin{tabular}{|l|c|c|}
  \hline
  Methods  & mIoU (\%) & FPS \\
  \hline\hline
  a) LGCNet & 69.5 & 124.1 \\
  b) LGCNet with ASPP  & 72.4 & 108.4  \\
  c) LGCNet with PPM & 72.5 & 114.1  \\
  d) LGCNet with MSCell  & 72.7 & 112.7  \\
  e) LGCNet with DCFC & \textbf{73.0} & 115.2  \\
  \hline
  \end{tabular}}
  }
  \end{center}
  \label{ppm}
  \end{table}
  
   We also conduct a ablation study for which features should be aggregated, and the result is shown in the Table \ref{tab:effi_aggregate}. Specifically, the cell $0$ is fixed for all experiment settings because there is only one cell with low level information (stride=4), and we explore different choices for other feature scales (i.e., stride=8 and stride=16). The numbers in the table is the average result of three repeated experiments with the same random seed on the Cityscapes validation set. The results show that cell $0$, cell $7$ and cell $13$ can achieve best trade-off between accuracy and latency comparison to other settings.

  \begin{table}[h]
    \caption{Ablation study for which features should be aggregated on Cityscapes validation dataset.}
    \begin{center}
    \setlength{\tabcolsep}{1mm}{
    \scalebox{1.1}{
    \begin{tabular}{|l|c|c|c|}
    \hline
    Cells to aggregate  & mIoU (\%) & FPS \\
    \hline\hline
   a) [0, 5, 13]  & 72.5 & 109.8 \\
   b) [0, 6, 13]  & 71.8 & 113.4 \\
   c) [0, 7, 13]  & \textbf{72.9} & 115.1 \\
   e) [0, 7, 10]  & 72.2 & 115.2 \\
   f) [0, 7, 11]  & 72.1 & 114.6 \\
   g) [0, 7, 12]  & 72.4 & 112.8 \\
    \hline
    \end{tabular}}
    }
    \end{center}
    \label{tab:effi_aggregate}
  \end{table}

  \paragraph{Effectiveness of the Latency Constraint}
  
\begin{figure}[h]
  \centering
  \includegraphics[width=3.6in]{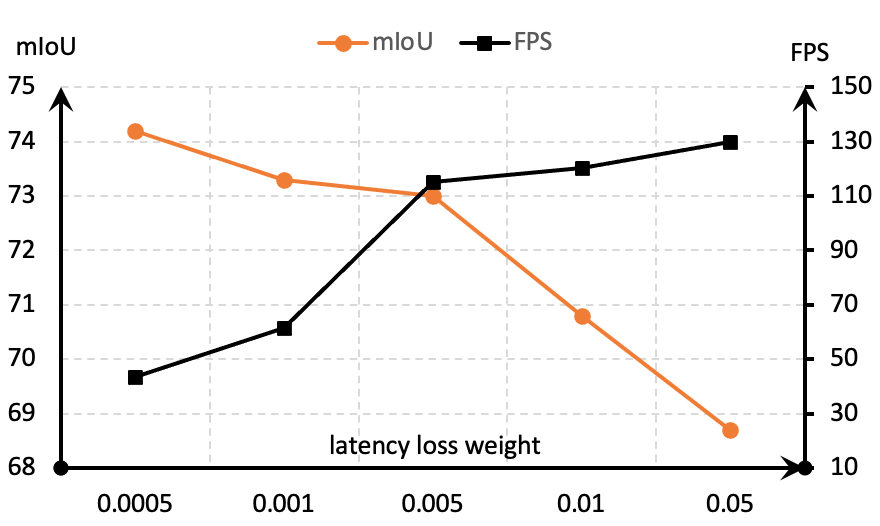}
  \caption{The accuracy on the Cityscapes validation set for different latency constraint settings.}
  \label{fig:compare}
  \end{figure}

  As mentioned above, LGCNet provides the ability to flexibly achieve a superior trade-off between performance and speed with latency-oriented optimization. We conducted a series of experiments with different loss weights $\beta$ in Equation \ref{equ:latency2}. Figure \ref{fig:compare} shows the variation of mIoU and latency as $\beta$ changes. With smaller $\beta$, we can obtain a model with higher accuracy, and vice-versa. When the $beta$ grows from 0.0005 to 0.005. When the $\beta$ increases from 0.0005 to 0.005, the latency dramatically decreases and performance somewhat degrades slightly. However, as the $\beta$ increases from 0.005 to 0.05, the performance declines rapidly while the latency decline is fairly limited. Thus in our experiments, we set $\beta$ as 0.005. It is obvious that the latency-oriented optimization works well for achieving a balance between accuracy and latency.

  \subsection{Network Visualization and Analysis}\label{visualization}
    \begin{figure*}
    \centering
    \includegraphics[width=6.0in]{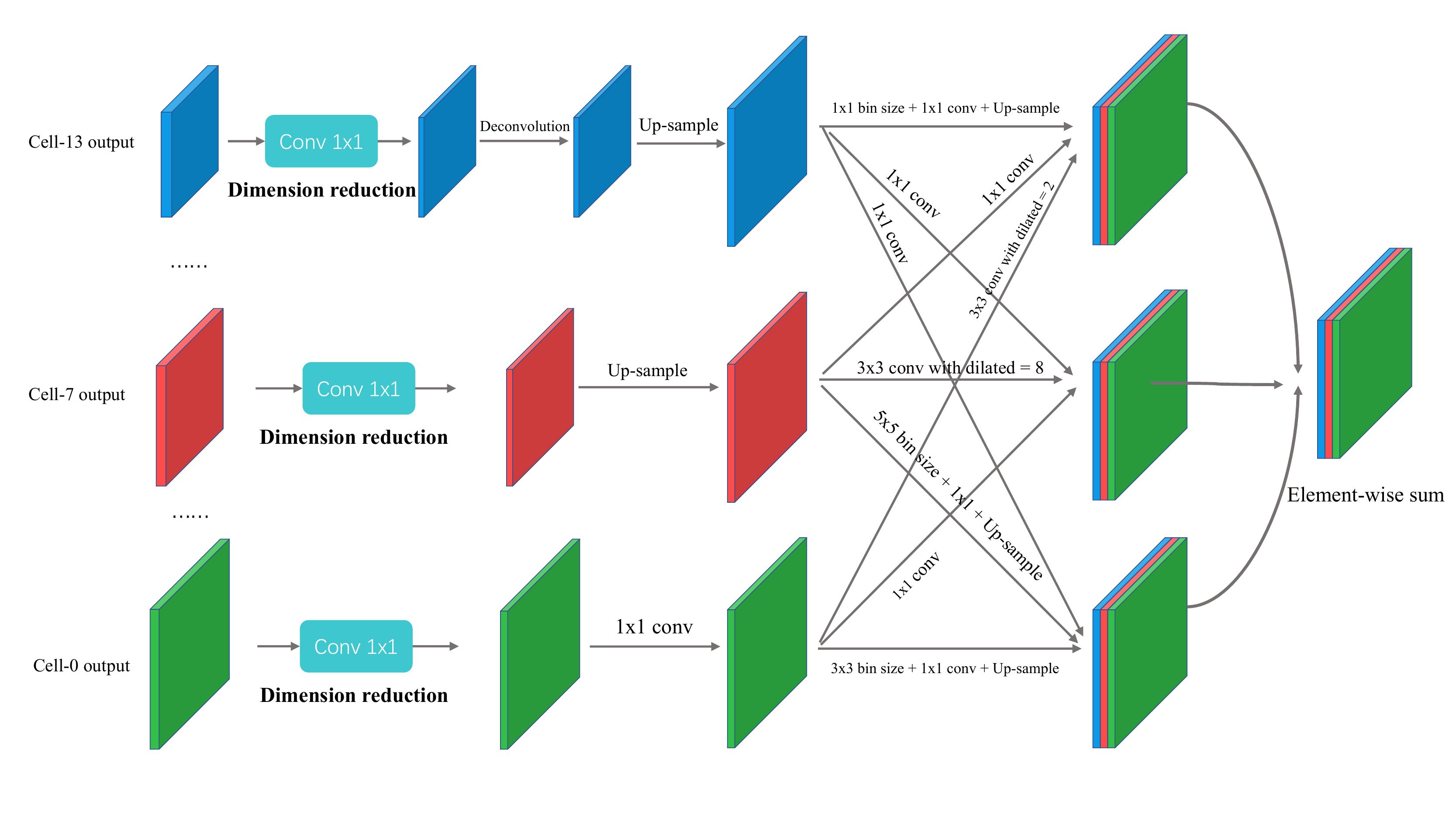}
    \caption{Illustration of the searched dense-connected fusion cell architecture.}
    \label{fig:searchedfusioncell}
    \end{figure*}
    
  \begin{figure*}
  \centering
  \includegraphics[width=7in]{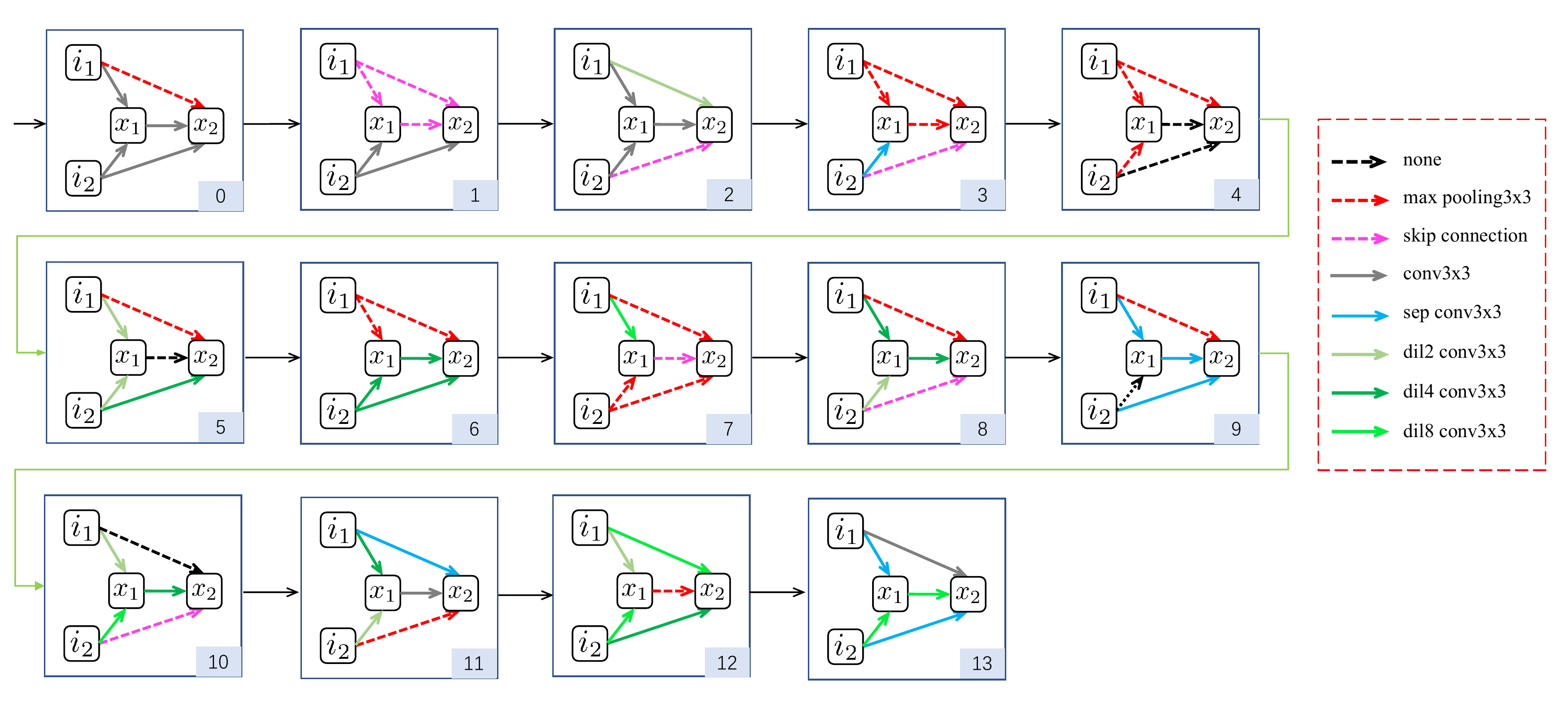}
  \caption{The network searched by our LGCNet with GGM exhibits the benefited property (e.g. more dilation convolution operations in deep layers and more low computational operations for fast speed) for real-time semantic segmentation.}
  \label{GDFAS_network}
  \end{figure*}
  
  \begin{figure*}
  \centering
  \includegraphics[width=7in]{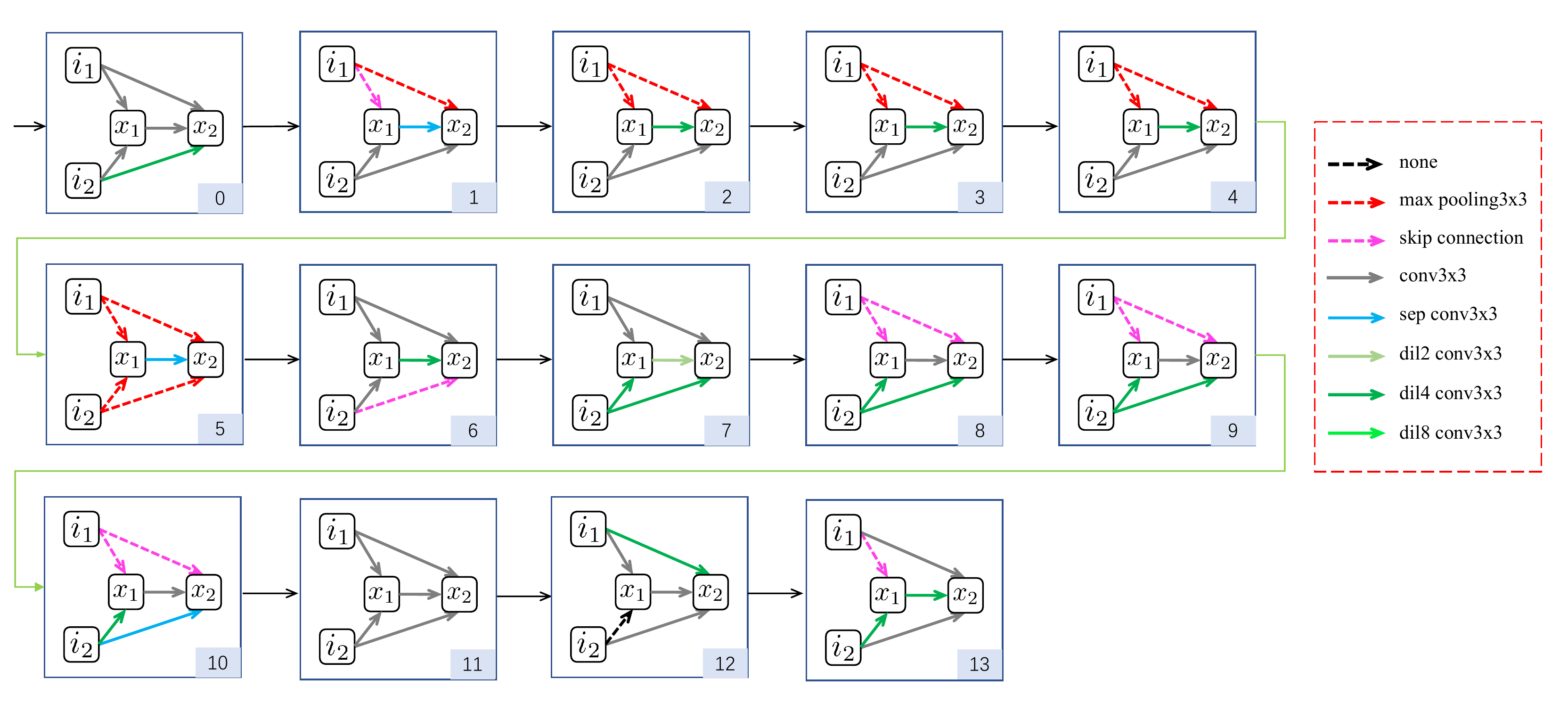}
  \caption{The network searched by our LGCNet with fully connected layer.}
  \label{fc_network}
  \end{figure*}
  
  \begin{figure*}
  \centering
  \includegraphics[width=7in]{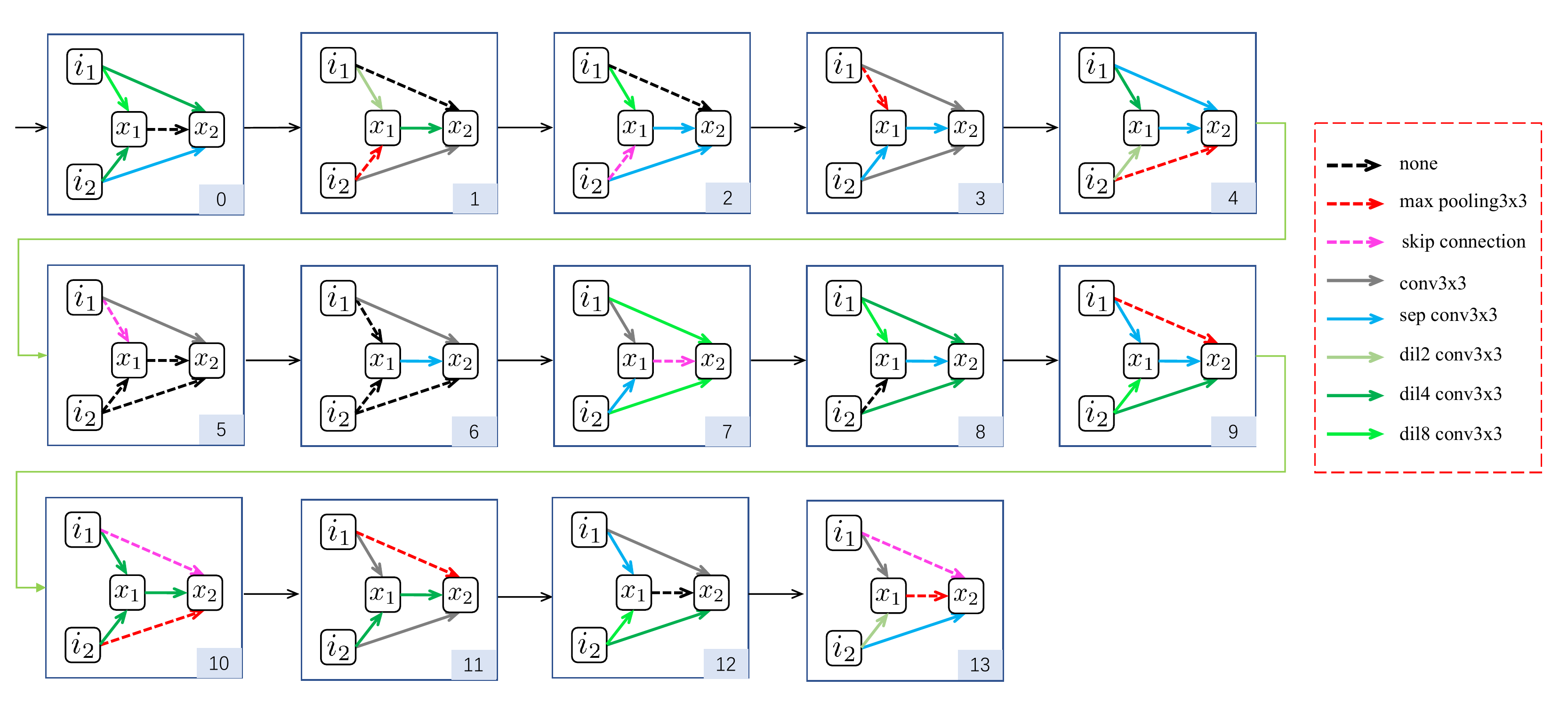}
  \caption{The randomly sampled network.}
  \label{random_network}
  \end{figure*}
  
  \paragraph{Network Visualization}
  As shown in Table \ref{tab:effi_gcn}, the network searched by our LGCNet with GGM has a smaller parameter size while achieving much higher performance. 
  The visualization results can effectively help to analyze which component brings in the performance improvement. 
  We thus visualize the networks searched on Cityscapes by the three methods: 1) LGCNet with GGM; 2) LGCNet with fully connected layer; and 3) random search in Figure \ref{GDFAS_network}, Figure \ref{fc_network} and Figure \ref{random_network}, respectively.
  
  Compared to the other methods, the network searched by our LGCNet with GGM shows the following three advantages:
  
  1) The cells in the shallow stage tend to choose light-weight operations (\ie, none, max pooling, skip connection), while the cells in the deep stage prefer the complicated ones, which is the goal of pursuing both high speed and performance as mentioned in the introduction of our study. Specifically, under the same latency loss weight, the network searched by our LGCNet with GGM comprises thirty light-weight operations (dashed-line arrow in the figure) with lower latency, whereas the other two methods utilize twenty-one and twenty-three light-weight operations, respectively. Our LGCNet with GGM, on the other hand, achieves superior performance, demonstrating the effectiveness of the concept of burden-sharing in a group of cells when they are aware of how much others are willing to contribute.
  
  2) The deeper layers tend to utilize larger receptive field operations (e.g. conv with dilation = 4 or 8), which plays a key role to improve performance in semantic segmentation \cite{chen2018deeplab,DBLP:journals/corr/ChenPSA17}. Specifically, the network searched by our LGCNet with GGM uses 11 large receptive field operations (denoted by the green arrow) in the last four cells and the other methods only use 4 or 8 operations, respectively.
  
  3) As we anticipated, the final structure searched by our LGCNet with GGM exhibits sufficient cell-level diversity. On the contrary, the network search by LGCNet with fully connected layer tends to employ similar structures, for instance, cell 7 is similar to cell 8 and 9, and cell 1 is similar to the cell 2, 3 and 4.

  Moreover, we also show the architecture of the searched dense-connected fusion cell in Figure \ref{fig:searchedfusioncell}. We clearly found that the selection of operations in the shallow stage (\ie low-level features) tends to choose more dilated convolution operations for achieving enough receptive field when performing the multi-scale feature fusion. As we expected, we obverse that the choice of dilated convolution operations in the low-level stage narrows the semantic gaps among the features from different scales.

  \paragraph{Analysis of the GCN-Guided Module}

  One concern is about what kind of role does GCN play in the search process. We suspect that its effectiveness is derived from the following two aspects: 1) to find a light-weight network, we do not allow the cell structures to share with each other to encourage structure diversity. Apparently, learning cells independently makes the search process much more difficult due to the enlarged search space and does not guarantee better performance, thus the GCN-Guided Module can be regraded as a regularization term to the search process. 2) We have discussed that $p(Z)$ is a fully factorizable joint distribution in the above section. As shown in Equation \ref{equ:softmax}, $p(Z_{h,i})$ for current cell becomes a conditional probability if the architecture parameter $\alpha_{h,i}$ depends on the probability $\alpha_{h,i}$ for previous cell. In this case, the GCN-Guided Module plays a role of modeling the condition in probability distribution $p(Z)$.

  \section{Conclusion}
  \par In this paper, a novel Graph-guided and Dense-Connected Fusion Architecture Search (LGCNet) framework is proposed to tackle the real-time semantic segmentation task. In contrast to the existing NAS approaches, which stack the same searched cell into a whole network, LGCNet explores to search different cell architectures and adopts the graph convolutional network to bridge the information connection among cells. In addition, we propose a novel dense-connected fusion cell that aggregates multi-level features in the network automatically to effectively fuse the low-level spatial details and high-level semantic context. Finally, a latency-oriented constraint is endowed into the search process for balancing accuracy and speed. Extensive experiments have demonstrated that LGCNet performs much better than the state-of-the-art real-time segmentation approaches.

\bibliographystyle{IEEEtran}
\bibliography{gas}

\begin{thebibliography}{10}
\providecommand{\url}[1]{#1}
\csname url@samestyle\endcsname
\providecommand{\newblock}{\relax}
\providecommand{\bibinfo}[2]{#2}
\providecommand{\BIBentrySTDinterwordspacing}{\spaceskip=0pt\relax}
\providecommand{\BIBentryALTinterwordstretchfactor}{4}
\providecommand{\BIBentryALTinterwordspacing}{\spaceskip=\fontdimen2\font plus
\BIBentryALTinterwordstretchfactor\fontdimen3\font minus
  \fontdimen4\font\relax}
\providecommand{\BIBforeignlanguage}[2]{{%
\expandafter\ifx\csname l@#1\endcsname\relax
\typeout{** WARNING: IEEEtran.bst: No hyphenation pattern has been}%
\typeout{** loaded for the language `#1'. Using the pattern for}%
\typeout{** the default language instead.}%
\else
\language=\csname l@#1\endcsname
\fi
#2}}
\providecommand{\BIBdecl}{\relax}
\BIBdecl

\bibitem{long2015fully}
J.~Long, E.~Shelhamer, and T.~Darrell, ``Fully convolutional networks for
  semantic segmentation,'' in \emph{CVPR}, 2015, pp. 3431--3440.

\bibitem{zhao2017pyramid}
H.~Zhao, J.~Shi, X.~Qi, X.~Wang, and J.~Jia, ``Pyramid scene parsing network,''
  in \emph{CVPR}, 2017, pp. 2881--2890.

\bibitem{chen2018deeplab}
L.-C. Chen, G.~Papandreou, I.~Kokkinos, K.~Murphy, and A.~L. Yuille, ``Deeplab:
  Semantic image segmentation with deep convolutional nets, atrous convolution,
  and fully connected crfs,'' \emph{IEEE trans. PAMI}, vol.~40, no.~4, pp.
  834--848, 2018.

\bibitem{DBLP:journals/corr/ChenPSA17}
L.~Chen, G.~Papandreou, F.~Schroff, and H.~Adam, ``Rethinking atrous
  convolution for semantic image segmentation,'' \emph{CoRR}, vol.
  abs/1706.05587, 2017.

\bibitem{DBLP:journals/tip/LiZCYTZX21}
X.~Li, L.~Zhang, G.~Cheng, K.~Yang, Y.~Tong, X.~Zhu, and T.~Xiang, ``Global
  aggregation then local distribution for scene parsing,'' \emph{{IEEE} Trans.
  Image Process.}, vol.~30, pp. 6829--6842, 2021.

\bibitem{DBLP:journals/tip/LiLYZCYTL21}
X.~Li, X.~Li, A.~You, L.~Zhang, G.~Cheng, K.~Yang, Y.~Tong, and Z.~Lin,
  ``Towards efficient scene understanding via squeeze reasoning,'' \emph{{IEEE}
  Trans. Image Process.}, vol.~30, pp. 7050--7063, 2021.

\bibitem{corr_SimonyanZ14a}
K.~Simonyan and A.~Zisserman, ``Very deep convolutional networks for
  large-scale image recognition,'' \emph{CoRR}, vol. abs/1409.1556, 2014.

\bibitem{conf_cvpr_HeZRS16}
K.~He, X.~Zhang, S.~Ren, and J.~Sun, ``Deep residual learning for image
  recognition,'' in \emph{CVPR}, 2016, pp. 770--778.

\bibitem{journals_corr_HuangLW16a}
G.~Huang, Z.~Liu, and K.~Q. Weinberger, ``Densely connected convolutional
  networks,'' \emph{CVPR}, pp. 1--9, 2016.

\bibitem{DBLP:conf/cvpr/Chollet17}
F.~Chollet, ``Xception: Deep learning with depthwise separable convolutions,''
  in \emph{CVPR}, 2017, pp. 1800--1807.

\bibitem{DBLP:journals/corr/abs-2107-13155}
X.~Li, H.~He, H.~Ding, K.~Yang, G.~Cheng, J.~Shi, and Y.~Tong, ``Improving
  video instance segmentation via temporal pyramid routing,'' \emph{CoRR}, vol.
  abs/2107.13155, 2021.

\bibitem{DBLP:conf/eccv/LiLZCSLTT20}
X.~Li, X.~Li, L.~Zhang, G.~Cheng, J.~Shi, Z.~Lin, S.~Tan, and Y.~Tong,
  ``Improving semantic segmentation via decoupled body and edge supervision,''
  in \emph{Computer Vision - {ECCV} 2020 - 16th European Conference, Glasgow,
  UK, August 23-28, 2020, Proceedings, Part {XVII}}, ser. Lecture Notes in
  Computer Science, A.~Vedaldi, H.~Bischof, T.~Brox, and J.~Frahm, Eds., vol.
  12362.\hskip 1em plus 0.5em minus 0.4em\relax Springer, 2020, pp. 435--452.

\bibitem{cordts2016cityscapes}
M.~Cordts, M.~Omran, S.~Ramos, T.~Rehfeld, M.~Enzweiler, R.~Benenson,
  U.~Franke, S.~Roth, and B.~Schiele, ``The cityscapes dataset for semantic
  urban scene understanding,'' in \emph{CVPR}, 2016, pp. 3213--3223.

\bibitem{everingham2015pascal}
M.~Everingham, S.~A. Eslami, L.~Van~Gool, C.~K. Williams, J.~Winn, and
  A.~Zisserman, ``The pascal visual object classes challenge: A
  retrospective,'' \emph{IJCV}, vol. 111, no.~1, pp. 98--136, 2015.

\bibitem{Camvid}
G.~J. Brostow, J.~Shotton, J.~Fauqueur, and R.~Cipolla, ``Segmentation and
  recognition using structure from motion point clouds,'' in \emph{ECCV (1)},
  2008, pp. 44--57.

\bibitem{zhao2018icnet}
H.~Zhao, X.~Qi, X.~Shen, J.~Shi, and J.~Jia, ``Icnet for real-time semantic
  segmentation on high-resolution images,'' in \emph{ECCV}, 2018, pp. 405--420.

\bibitem{badrinarayanan2017segnet}
V.~Badrinarayanan, A.~Kendall, and R.~Cipolla, ``Segnet: A deep convolutional
  encoder-decoder architecture for image segmentation,'' \emph{IEEE trans.
  PAMI}, vol.~39, no.~12, pp. 2481--2495, 2017.

\bibitem{paszke2016enet}
A.~Paszke, A.~Chaurasia, S.~Kim, and E.~Culurciello, ``Enet: A deep neural
  network architecture for real-time semantic segmentation,''
  \emph{arXiv:1606.02147}, 2016.

\bibitem{DBLP:conf/eccv/YuWPGYS18}
C.~Yu, J.~Wang, C.~Peng, C.~Gao, G.~Yu, and N.~Sang, ``Bisenet: Bilateral
  segmentation network for real-time semantic segmentation,'' in \emph{ECCV},
  2018, pp. 334--349.

\bibitem{yu2021bisenet}
C.~Yu, C.~Gao, J.~Wang, G.~Yu, C.~Shen, and N.~Sang, ``Bisenet v2: Bilateral
  network with guided aggregation for real-time semantic segmentation,''
  \emph{International Journal of Computer Vision}, pp. 1--18, 2021.

\bibitem{li2019dfanet}
H.~Li, P.~Xiong, H.~Fan, and J.~Sun, ``Dfanet: Deep feature aggregation for
  real-time semantic segmentation,'' in \emph{CVPR}, 2019, pp. 9522--9531.

\bibitem{DBLP:conf/iclr/LiuSY19}
H.~Liu, K.~Simonyan, and Y.~Yang, ``{DARTS:} differentiable architecture
  search,'' in \emph{ICLR}, 2019.

\bibitem{zoph2018learning}
B.~Zoph, V.~Vasudevan, J.~Shlens, and Q.~V. Le, ``Learning transferable
  architectures for scalable image recognition,'' in \emph{CVPR}, 2018, pp.
  8697--8710.

\bibitem{negrinho2017deeparchitect}
R.~Negrinho and G.~Gordon, ``Deeparchitect: Automatically designing and
  training deep architectures,'' \emph{arXiv:1704.08792}, 2017.

\bibitem{krause2017dynamic}
B.~Krause, E.~Kahembwe, I.~Murray, and S.~Renals, ``Dynamic evaluation of
  neural sequence models,'' \emph{arXiv:1709.07432}, 2017.

\bibitem{DBLP:conf/icml/PhamGZLD18}
H.~Pham, M.~Y. Guan, B.~Zoph, Q.~V. Le, and J.~Dean, ``Efficient neural
  architecture search via parameter sharing,'' in \emph{ICML}, 2018, pp.
  4092--4101.

\bibitem{cai2018proxylessnas}
H.~Cai, L.~Zhu, and S.~Han, ``Proxylessnas: Direct neural architecture search
  on target task and hardware,'' \emph{arXiv:1812.00332}, 2018.

\bibitem{DBLP:conf/iclr/XieZLL19}
S.~Xie, H.~Zheng, C.~Liu, and L.~Lin, ``{SNAS:} stochastic neural architecture
  search,'' in \emph{ICLR}, 2019.

\bibitem{nekrasov2019fast}
V.~Nekrasov, H.~Chen, C.~Shen, and I.~Reid, ``Fast neural architecture search
  of compact semantic segmentation models via auxiliary cells,'' in
  \emph{Proceedings of the IEEE/CVF Conference on Computer Vision and Pattern
  Recognition}, 2019, pp. 9126--9135.

\bibitem{nirkin2021hyperseg}
Y.~Nirkin, L.~Wolf, and T.~Hassner, ``Hyperseg: Patch-wise hypernetwork for
  real-time semantic segmentation,'' in \emph{Proceedings of the IEEE/CVF
  Conference on Computer Vision and Pattern Recognition}, 2021, pp. 4061--4070.

\bibitem{DBLP:conf/eccv/ChenZPSA18}
L.~Chen, Y.~Zhu, G.~Papandreou, F.~Schroff, and H.~Adam, ``Encoder-decoder with
  atrous separable convolution for semantic image segmentation,'' in
  \emph{ECCV}, 2018, pp. 833--851.

\bibitem{zhang2019customizable}
Y.~Zhang, Z.~Qiu, J.~Liu, T.~Yao, D.~Liu, and T.~Mei, ``Customizable
  architecture search for semantic segmentation,'' in \emph{CVPR}, 2019, pp.
  11\,641--11\,650.

\bibitem{wu2019fbnet}
B.~Wu, X.~Dai, P.~Zhang, Y.~Wang, F.~Sun, Y.~Wu, Y.~Tian, P.~Vajda, Y.~Jia, and
  K.~Keutzer, ``Fbnet: Hardware-aware efficient convnet design via
  differentiable neural architecture search,'' in \emph{CVPR}, 2019, pp.
  10\,734--10\,742.

\bibitem{2019_SqueezeNAS}
A.~Shaw, D.~Hunter, F.~Iandola, and S.~Sidhu, ``{SqueezeNAS}: Fast neural
  architecture search for faster semantic segmentation,'' in \emph{ICCV Neural
  Architects Workshop}, 2019.

\bibitem{howard2017mobilenets}
A.~G. Howard, M.~Zhu, B.~Chen, D.~Kalenichenko, W.~Wang, T.~Weyand,
  M.~Andreetto, and H.~Adam, ``Mobilenets: Efficient convolutional neural
  networks for mobile vision applications,'' \emph{arXiv:1704.04861}, 2017.

\bibitem{sun2021real}
P.~Sun, J.~Wu, S.~Li, P.~Lin, J.~Huang, and X.~Li, ``Real-time semantic
  segmentation via auto depth, downsampling joint decision and feature
  aggregation,'' \emph{International Journal of Computer Vision}, vol. 129,
  no.~5, pp. 1506--1525, 2021.

\bibitem{minsky1988society}
M.~Minsky, \emph{The Society of Mind}.\hskip 1em plus 0.5em minus 0.4em\relax
  Simon \& Schuster, 1988.

\bibitem{kipf2016semi}
T.~N. Kipf and M.~Welling, ``Semi-supervised classification with graph
  convolutional networks,'' \emph{arXiv:1609.02907}, 2016.

\bibitem{gas}
P.~Lin, P.~Sun, G.~Cheng, S.~Xie, X.~Li, and J.~Shi, ``Graph-guided
  architecture search for real-time semantic segmentation,'' in \emph{CVPR, to
  appear}, 2020.

\bibitem{chen2017deeplab}
L.-C. Chen, G.~Papandreou, I.~Kokkinos, K.~Murphy, and A.~L. Yuille, ``Deeplab:
  Semantic image segmentation with deep convolutional nets, atrous convolution,
  and fully connected crfs,'' \emph{IEEE transactions on pattern analysis and
  machine intelligence}, vol.~40, no.~4, pp. 834--848, 2017.

\bibitem{DBLP:conf/iclr/ZophL17}
B.~Zoph and Q.~V. Le, ``Neural architecture search with reinforcement
  learning,'' in \emph{ICLR}, 2017.

\bibitem{DBLP:journals/corr/abs-1812-05285}
M.~Guo, Z.~Zhong, W.~Wu, D.~Lin, and J.~Yan, ``{IRLAS:} inverse reinforcement
  learning for architecture search,'' \emph{CoRR}, vol. abs/1812.05285, 2018.

\bibitem{DBLP:journals/corr/abs-1802-01548}
\BIBentryALTinterwordspacing
E.~Real, A.~Aggarwal, Y.~Huang, and Q.~V. Le, ``Regularized evolution for image
  classifier architecture search,'' \emph{CoRR}, vol. abs/1802.01548, 2018.
  [Online]. Available: \url{http://arxiv.org/abs/1802.01548}
\BIBentrySTDinterwordspacing

\bibitem{DBLP:journals/corr/abs-1808-00193}
\BIBentryALTinterwordspacing
Y.~Chen, Q.~Zhang, C.~Huang, L.~Mu, G.~Meng, and X.~Wang, ``Reinforced
  evolutionary neural architecture search,'' \emph{CoRR}, vol. abs/1808.00193,
  2018. [Online]. Available: \url{http://arxiv.org/abs/1808.00193}
\BIBentrySTDinterwordspacing

\bibitem{bender2018understanding}
G.~Bender, P.-J. Kindermans, B.~Zoph, V.~Vasudevan, and Q.~Le, ``Understanding
  and simplifying one-shot architecture search,'' in \emph{ICML}, 2018, pp.
  549--558.

\bibitem{brock2017smash}
A.~Brock, T.~Lim, J.~M. Ritchie, and N.~Weston, ``Smash: one-shot model
  architecture search through hypernetworks,'' \emph{arXiv:1708.05344}, 2017.

\bibitem{chen2019progressive}
X.~Chen, L.~Xie, J.~Wu, and Q.~Tian, ``Progressive differentiable architecture
  search: Bridging the depth gap between search and evaluation,''
  \emph{arXiv:1904.12760}, 2019.

\bibitem{noy2019asap}
A.~Noy, N.~Nayman, T.~Ridnik, N.~Zamir, S.~Doveh, I.~Friedman, R.~Giryes, and
  L.~Zelnik-Manor, ``Asap: Architecture search, anneal and prune,''
  \emph{arXiv:1904.04123}, 2019.

\bibitem{tan2019mnasnet}
M.~Tan, B.~Chen, R.~Pang, V.~Vasudevan, M.~Sandler, A.~Howard, and Q.~V. Le,
  ``Mnasnet: Platform-aware neural architecture search for mobile,'' in
  \emph{CVPR}, 2019, pp. 2820--2828.

\bibitem{DBLP:conf/nips/ChenCZPZSAS18}
L.~Chen, M.~D. Collins, Y.~Zhu, G.~Papandreou, B.~Zoph, F.~Schroff, H.~Adam,
  and J.~Shlens, ``Searching for efficient multi-scale architectures for dense
  image prediction,'' in \emph{NeurIPS}, 2018, pp. 8713--8724.

\bibitem{DBLP:journals/corr/abs-1901-02985}
C.~Liu, L.-C. Chen, F.~Schroff, H.~Adam, W.~Hua, A.~Yuille, and L.~Fei-Fei,
  ``Auto-deeplab: Hierarchical neural architecture search for semantic image
  segmentation,'' in \emph{CVPR}, 2019.

\bibitem{wang2018videos}
X.~Wang and A.~Gupta, ``Videos as space-time region graphs,'' in \emph{ECCV},
  2018, pp. 399--417.

\bibitem{stgcn2018aaai}
S.~Yan, Y.~Xiong, and D.~Lin, ``Spatial temporal graph convolutional networks
  for skeleton-based action recognition,'' in \emph{AAAI}, 2018.

\bibitem{Zhang2019GraphHF}
C.~Zhang, M.~Ren, and R.~Urtasun, ``Graph hypernetworks for neural architecture
  search,'' \emph{ICLR}, vol. abs/1810.05749, 2019.

\bibitem{maddison2016concrete}
C.~J. Maddison, A.~Mnih, and Y.~W. Teh, ``The concrete distribution: A
  continuous relaxation of discrete random variables,''
  \emph{arXiv:1611.00712}, 2016.

\bibitem{pytorch}
A.~Paszke, S.~Gross, F.~Massa, A.~Lerer, J.~Bradbury, G.~Chanan, T.~Killeen,
  Z.~Lin, N.~Gimelshein, L.~Antiga, A.~Desmaison, A.~Kopf, E.~Yang, Z.~DeVito,
  M.~Raison, A.~Tejani, S.~Chilamkurthy, B.~Steiner, L.~Fang, J.~Bai, and
  S.~Chintala, ``Pytorch: An imperative style, high-performance deep learning
  library,'' in \emph{NIPS}, 2019.

\bibitem{deng2009imagenet}
J.~Deng, W.~Dong, R.~Socher, L.-J. Li, K.~Li, and L.~Fei-Fei, ``Imagenet: A
  large-scale hierarchical image database,'' in \emph{CVPR}.\hskip 1em plus
  0.5em minus 0.4em\relax IEEE, 2009, pp. 248--255.

\bibitem{DBLP:journals/corr/WuSH16a}
Z.~Wu, C.~Shen, and A.~van~den Hengel, ``High-performance semantic segmentation
  using very deep fully convolutional networks,'' \emph{CoRR}, vol.
  abs/1604.04339, 2016.

\bibitem{treml2016speedingSQ}
M.~Treml, J.~Arjona-Medina, T.~Unterthiner, R.~Durgesh, F.~Friedmann,
  P.~Schuberth, A.~Mayr, M.~Heusel, M.~Hofmarcher, M.~Widrich \emph{et~al.},
  ``Speeding up semantic segmentation for autonomous driving,'' in \emph{MLITS,
  NIPS Workshop}, vol.~2, 2016, p.~7.

\bibitem{SwiftNet}
M.~Orsic, I.~Kreso, P.~Bevandic, and S.~Segvic, ``In defense of pre-trained
  imagenet architectures for real-time semantic segmentation of road-driving
  images,'' \emph{CVPR}, pp. 12\,599--12\,608, 2019.

\bibitem{li2019partialdfnet}
X.~Li, Y.~Zhou, Z.~Pan, and J.~Feng, ``Partial order pruning: for best
  speed/accuracy trade-off in neural architecture search,'' in
  \emph{Proceedings of IEEE Conference on Computer Vision and Pattern
  Recognition (CVPR)}, 2019.

\bibitem{mehta2018espnet}
S.~Mehta, M.~Rastegari, A.~Caspi, L.~Shapiro, and H.~Hajishirzi, ``Espnet:
  Efficient spatial pyramid of dilated convolutions for semantic
  segmentation,'' in \emph{ECCV}, 2018, pp. 552--568.

\bibitem{Mehta2019ESPNetv2AL}
S.~Mehta, M.~Rastegari, L.~Shapiro, and H.~Hajishirzi, ``Espnetv2: A
  light-weight, power efficient, and general purpose convolutional neural
  network,'' \emph{2019 IEEE/CVF Conference on Computer Vision and Pattern
  Recognition (CVPR)}, pp. 9182--9192, 2019.

\bibitem{Romera2018ERFNetER}
E.~Romera, J.~M. {\'A}lvarez, L.~M. Bergasa, and R.~Arroyo, ``Erfnet: Efficient
  residual factorized convnet for real-time semantic segmentation,'' \emph{IEEE
  Transactions on Intelligent Transportation Systems}, vol.~19, pp. 263--272,
  2018.

\bibitem{Zhuang2019ShelfNetFF}
J.~Zhuang, J.~Yang, L.~Gu, and N.~Dvornek, ``Shelfnet for fast semantic
  segmentation,'' \emph{2019 IEEE/CVF International Conference on Computer
  Vision Workshop (ICCVW)}, pp. 847--856, 2019.

\bibitem{Chen2020FasterSegSF}
W.~Chen, X.~Gong, X.~Liu, Q.~Zhang, Y.~Li, and Z.~Wang, ``Fasterseg: Searching
  for faster real-time semantic segmentation,'' \emph{ICLR}, vol.
  abs/1912.10917, 2020.

\bibitem{fan2021rethinking}
M.~Fan, S.~Lai, J.~Huang, X.~Wei, Z.~Chai, J.~Luo, and X.~Wei, ``Rethinking
  bisenet for real-time semantic segmentation,'' in \emph{Proceedings of the
  IEEE/CVF Conference on Computer Vision and Pattern Recognition}, 2021, pp.
  9716--9725.

\bibitem{DBLP:conf/uai/LiT19}
L.~Li and A.~Talwalkar, ``Random search and reproducibility for neural
  architecture search,'' in \emph{Proceedings of the Thirty-Fifth Conference on
  Uncertainty in Artificial Intelligence, {UAI} 2019, Tel Aviv, Israel, July
  22-25, 2019}, 2019, p. 129.

\end{thebibliography}

\end{document}